\documentclass{article}

\PassOptionsToPackage{numbers}{natbib}
\usepackage[preprint]{neurips_2020}

\usepackage{multibib}

\usepackage[utf8]{inputenc} 
\usepackage[T1]{fontenc}    
\usepackage{hyperref}       
\usepackage{url}            
\usepackage{booktabs}       
\usepackage{amsfonts}       
\usepackage{nicefrac}       
\usepackage{microtype}      
\usepackage{color}
\usepackage{appendix}
\usepackage{subcaption}
\usepackage{mathtools}

\usepackage{epsfig}
\usepackage{graphicx}
\usepackage{amsmath}
\usepackage{amssymb}
\usepackage{amsthm}
\usepackage{verbatim}

\usepackage{multicol}

\usepackage{algorithm,algorithmic}
\renewcommand{\algorithmiccomment}[1]{\bgroup\hfill\footnotesize//~#1\egroup}
\newcommand{\xxx}{\mbox{\textit{HDP-VFL}}}

\def\trans{^{\rm T}}

\def\calL{\mathcal{L}}
\def\calS{\mathcal{S}}

\def\X{\mathbf{X}}
\def\Z{\mathbf{Z}}

\def\e{\mathbf{e}}
\def\s{\mathbf{s}}

\def\g{\mathbf{g}}

\def\x{\mathbf{x}}

\def\y{\mathbf{y}}

\def\w{\mathbf{w}}

\newtheorem{theorem}{Theorem}

\newtheorem{lemma}{Lemma}
\newtheorem{definition}{Definition}

\newtheorem{property}{Property}

\definecolor{Gray}{gray}{0.85}
\definecolor{LightCyan}{rgb}{0.88,1,1}

\title{Hybrid Differentially Private Federated Learning on Vertically Partitioned Data}

\author{%
	Chang Wang \\
	Tencent Inc. \\
	Beijing, China\\
	\texttt{coracwang@tencent.com} \\
	\And
	Jian Liang \\
	Tencent Inc. \\
	Beijing, China \\
	\texttt{joshualiang@tencent.com} \\
	\AND
	Mingkai Huang \\
	Tencent Inc. \\
	Beijing, China  \\
	\texttt{mingkhuang@tencent.com} \\
	\And
	Bing Bai \\
	Tencent Inc. \\
	Beijing, China \\
	\texttt{icebai@tencent.com} \\
	\And
	Kun Bai \\
	Tencent Inc. \\
	Guangzhou, China \\
	\texttt{kunbai@tencent.com} \\
	\And
	Hao Li\thanks{Corresponding Author.} \\
	Tencent Inc. \\
	Beijing, China \\
	\texttt{leehaoli@tencent.com} \\
}

\begin{document}
	
	\maketitle
	
	\begin{abstract}
		We present \xxx, the first hybrid differentially private (DP) framework for vertical federated learning (VFL) to demonstrate that it is possible to jointly learn a generalized linear model (GLM) from vertically partitioned data with only a negligible cost, w.r.t. training time, accuracy, etc., comparing to idealized non-private VFL. Our work builds on the recent advances in VFL-based collaborative training among different organizations which rely on protocols like Homomorphic Encryption (HE) and Secure Multi-Party Computation (MPC) to secure computation and training. In particular, we analyze how VFL's intermediate result (IR) can leak private information of the training data during communication and design a DP-based privacy-preserving algorithm to ensure the data confidentiality of VFL participants. We mathematically prove that our algorithm not only provides utility guarantees for VFL, but also offers multi-level privacy, i.e. DP w.r.t. IR and joint differential privacy (JDP) w.r.t. model weights. Experimental results demonstrate that our work, under adequate privacy budgets, is quantitatively and qualitatively similar to GLMs, learned in idealized non-private VFL setting, rather than the increased cost in memory and processing time in most prior works based on HE or MPC. Our codes will be released if this paper is accepted.
	\end{abstract}
	
	\section{Introduction}
	Vertical federated learning (VFL)\cite{vfl} has been recognized as one of the effective solutions for encouraging enterprise-level data collaborations while respecting data privacy\cite{fl_concepts}, required by the strict  government regulations like Europe's General Data Privacy Regulations (GDPR\cite{10.5555/3152676}). Unlike horizontal federated learning (HFL)\cite{hfl, fl_sys_design} setting in which the decentralized datasets share the same feature space but little intersection on the sample space, in VFL setting, the datasets of different organizations share the same or similar sample space but differ in feature space. Therefore the VFL participants need to jointly learn a model together\cite{vfl}, rather than independently learn models from their local data like normal HFL participants do.
	
	The difference in data distribution leads to different focus on data protection in HFL and VFL. In HFL, gradients, trained with each participant's local data and vulnerable to information leakage\cite{dl_ahe}, are sent from each participant to server for a secure aggregation\cite{secure_aggregation}. The numerous participants of HFL are mostly personal smart phones or edge devices with limited computation power and network bandwidth, thus the goal of gradient protection in HFL setting is mainly achieved by differential privacy (DP)\cite{dwork2006calibrating, abadi2016deep, cpsgd, dp_fedavg, bayesian_dp}, secret sharing\cite{secure_aggregation}, and so on. In VFL setting, however, few enterprise-level participants jointly learn a machine learning model with their own data by merely exchanging intermediate result (IR), e.g. scalar inner product\cite{vfl} in generalized linear model (GLM). Since IRs might leak training data information\cite{abadi2016deep, chaudhuri2011differentially}, they are mostly protected by additively homomorphic encryption (HE)\cite{vfl, dl_ahe} and secure multi-party computation (MPC)\cite{secureml} in existing VFL, thanks to the sufficient computation power and network bandwidth of enterprise-level participants. In addition, given by the similar settings and assumptions, most existing VFL algorithms follow traditional privacy-preserving multi-party machine learning methods\cite{fl_concepts, lr_he, lr_he_scale, aono2016scalable, chaudhuri2008privacy}, by applying Taylor approximation to the loss functions, so that HE can be adopted to protect the calculation of polynomial tasks in VFL's joint training.
	
	We observe several drawbacks of VFL's data protection using HE. We measure large overhead on memory cost and processing time with HE, similar to the results in \cite{kim2018secure}. For example, a VFL-based logistic regression task takes few minutes to finish training if IR is not securely computed and transmitted, while an HE version of VFL-based logistic regression takes hours. In addition, as mentioned in \cite{fl_concepts}, most existing VFL methods require a third-party collaborator to ensure data confidentiality during training process. Moreover, It is non-trivial task to approximate certain critical functions, e.g., loss function in machine learning models using only low-degree polynomials before HE, and naive approximation may lead to big errors and makes the solutions intractable\cite{kim2018secure}. Although many research efforts have been devoted to gradient protection with DP in HFL, surprisingly, we find no prior work on protecting VFL's data confidentiality using DP. Therefore, we are motivated to propose the first differentially private framework to enforce the data confidentiality of VFL participants with negligible cost, in terms of training time, accuracy, and so on. 
	
	The contribution of this paper is threefold. First, to the best of our knowledge, we present \xxx, the first differentially private framework for VFL. By thoroughly analyzing the sensitivity of VFL's IR and conducting perturbation of IR directly within each training iteration among VFL's participants, our method doesn't need to perform Taylor approximation to the loss function, and meanwhile no HE is required, thus \xxx~could greatly boost VFL's performance. Second, we mathematically prove that \xxx~not only provides utility guarantees for VFL, but also offers multi-level privacy, i.e. DP w.r.t. IR and JDP w.r.t. model weights, for VFL's data protection. Third, by not relying on any third-party collaborator to ensure data confidentiality, \xxx~is easy to deploy.
	
	\section{Related Works}
	Although the local raw data is not exposed in FL setting, FL on its own still lacks theoretical privacy guarantees\cite{bayesian_dp}, and may leak sensitive information about the training data\cite{fl_concepts}. 
	Therefore, the combination of FL and proper privacy-preserving mechanisms, such as DP\cite{dwork2006calibrating}, HE\cite{he1978}, MPC\cite{smc1987}, etc., is a necessity to alleviate FL's privacy risks.
	
	\textbf{Privacy-preserving HFL:}
	Most privacy-preserving HFL systems are realized based on DP, MPC, and encryption, due to limited computation power and network bandwidth\cite{secure_aggregation}. For example, \citet{secure_aggregation} proposed a secure aggregation scheme based on MPC to allow server to obtain an aggregation result without learning data information of each participant. \citet{cpsgd} proposed cpSGD, a communication-efficient DP mechanism using binomial noise to avoid floating point representation issues. \citet{dp_fedavg} proposed DP-FedAvg, a differentially private version of vanilla FedAvg. \citet{bayesian_dp} proposed Bayesian differential privacy, a relaxation of DP for FL with a tighter privacy budget so that FL task over population with similarly distributed data could converge faster than DP-FedAvg. Unlike the existing methods providing gradient-level perturbation, our method focuses on IR  perturbation within each multi-party SGD iteration, which is unique in VFL.
	
	\textbf{Privacy-preserving VFL:}
	Unlike HFL releases summative private information (e.g. averaged gradients) w.r.t. some data instances, VFL releases summative private information (e.g. inner-products between data and parameters as scalar IR) w.r.t. some dimensions, which requires unique privacy-preserving solutions. With sufficient computation power and network bandwidth, most privacy-preserving VFL systems adopted time-consuming and memory-consuming\cite{kim2018secure} HE or MPC to protect the IR during joint training\cite{vfl, secureboost, fed_transfer} to pursue models with lossless prediction performance, which was assumed to be hard for DP\cite{vfl} although DP were dominate in traditional research on privacy-preserving machine learning on vertically partitioned data\cite{dwork2004privacy-preserving, approximate_ppdm}. Unlike existing privacy-preserving VFL, our method \xxx~proposes using DP to protect the training data of VFL participants. In addition, we mathematically prove \xxx's multi-level privacy and utility guarantees. 
	

	\section{Preliminaries}
	\label{preliminaries}
	This section reviews key definitions. 
	
	\textbf{Vertical Federated Learning (VFL).}
	VFL is applicable to the cases that several datasets, owned by various enterprise-level parties, share the same or similar sample space, i.e., sample IDs, but differ in feature space. Besides, only the party launching a specific joint training task owns the target vector. We define the party with target vector as the ``\textbf{active party}'' and the others as the ``\textbf{passive party}''.

	We denote VFL's datasets as $D^{m} = (\X^m, \y) = \{(\X^1), \ldots,(\X^i,\y), \ldots, (\X^m)\}$, where $\X^i\in \mathbb{R}^{n \times d_i}$  is the data matrix of the $i$-th party,
	and $\y \in \mathbb{R}^{n \times 1} $ is the target vector held by active party. When a specific VFL task only involves one active party and one passive party, 
	we also denote active party's data as  $(\X^{A},\y^{A})$ and passive party's data as $(\X^{B})$ within this paper. 
	Our goal is to support VFL-based model joint training privately and efficiently, and herein we take 
	generalized linear model (GLM) as an example. We define a two-party 
	VFL-based objective function as:
	
	\begin{equation}
	\label{equ:obj-function}
	\widehat{\w} = \mathop{\arg\min}_{\w} \calL(\w)= \frac{1}{n}\sum_{i = 1}^{n}\ell(\theta_i, \y_i) + \lambda g(\w), \ \mbox{s.t.}  \ \theta_i =\x_i\w, \forall i.
	\end{equation}
	
	where $n$ is the number of common entities after VFL's entity-resolution protocol \cite{vfl}, $\x_i = (\x_i^{A}, \x_i^{B})$, $\w = (\w^{A}, \w^{B})$, and $\y_i = \y_i^{A}$. 
	$\w \in \mathbb{R}^{d}$ is the vector of model weights, $\calL$ is the objective function, $\ell$ is the loss function for each data sample, and $\theta_i$ is the natural parameter for sample $i$. The $g(\cdot)$ is a regularization term, such as $\ell_1$ or $\ell_2$ regularization.
	To make sure the raw data $\x_i^{A}$ and $\x_i^{B}$, and target vector $\y_i^{A}$, 
	are not exposed to each other, meanwhile gradient and loss calculation are still possible at both parties, the secure version of \emph{intermediate result (IR)}, denoted as $Sec[\cdot]$, needs to be exchanged between VFL participants in each SGD iteration. 
	Currently existing $Sec[\cdot]$ in VFL is based on HE \citep{vfl, secureboost, fed_transfer}, and this paper presents a DP-based solution.
	
	\textbf{Differential Privacy (DP).} DP is concerned with whether the output of a computation over a dataset can leak information about individual entries in the dataset. To prevent leakage, \emph{randomness} is introduced into the computation to hide details of individual entries.
	
	\begin{definition}[Differential Privacy \cite{dwork2014algorithmic}]\label{df:dp}
		A randomized algorithm $\mathcal{A}: \mathcal{D} \rightarrow \mathcal{R}$ with domain $\mathcal{D}$ and range $\mathcal{R}$ satisfies $(\epsilon, \delta)$-differential privacy if for any two adjacent datasets $D, D' \in \mathcal{D}$ that differ by a single data instance and
		for any set of outcomes $\mathcal{S} \subset \mathcal{R}$, the following holds:
		\begin{align*}
		\mathbb{P}[\mathcal{A}(D)\in \mathcal{S}]
		\leq \exp(\epsilon)\mathbb{P}[\mathcal{A}(D')\in \mathcal{S}] + \delta.
		\end{align*}
	\end{definition}
	
	The privacy loss pair $(\epsilon,\delta)$ is referred to as the privacy budget/loss, and it quantifies the privacy risk of algorithm $\mathcal{A}$. The intuition is that it is difficult for a potential attacker to infer whether a certain data point has been changed in, or added into, the input $D$ based on a change in the output distribution. Consequently, the information of any single data point is protected. In our VFL setting, for active party and passive party, ($\x_i^A, y_i^A$) and $\x_i^B$  are treated as a ``single entry'' by Definition \ref{df:dp}, respectively.

	\begin{definition}[Joint Differential Privacy \cite{kearns2014mechanism}]\label{df:jdp}
		A randomized mechanism $\mathcal{M}: \mathcal{D} \rightarrow \mathcal{R}$ whose output is an $n$-tuple satisfies $(\epsilon,\delta)$-joint differential privacy if for any party $i\in \{1,2,\cdots,m\}$, any two adjacent datasets $D_i, D_i'$ of party $i$ that differ by a single data instance, all inputs $D_{-i}$ from any other parties except for party $i$, and any set of outcomes  $\mathcal{S} \subset \mathcal{R}^{n-1}$, the following holds:
		\begin{align*}
		\mathbb{P}[\mathcal{M}(D_i;D_{-i})_{-i}\in \mathcal{S}]
		\leq \exp(\epsilon)\mathbb{P}[\mathcal{M}({D_i}';D_{-i})_{-i}\in \mathcal{S}] + \delta.
		\end{align*}
	\end{definition}
	The privacy loss pair $(\epsilon,\delta)$ is referred to as the privacy budget/loss, and it quantifies the privacy risk of mechanism $\mathcal{M}$. 
	

	\begin{definition}[Sensitivity \cite{dwork2014algorithmic}]
		The sensitivity of a function $f: D \rightarrow \mathbb{R}^d$ is defined as:
		\begin{align*}
		\Delta_2(f) = \max_{D, D'}\|f(D) -f(D') \|,
		\end{align*}
		for all datasets $D$ and $D'$ that differ by at most one instance, where $\|\cdot \|$ { is specified by a particular mechanism. For example, the Gaussian mechanism~\citep{dwork2014analyze} requires the $\ell_2$ norm, and the Laplace mechanism~\citep{dwork2014algorithmic} requires the $\ell_1$ norm.}
	\end{definition}
	
	In this paper, we adopt the Gaussian mechanism for flexible usage.
	
	\begin{lemma}[Gaussian Mechanism \cite{dwork2014analyze}]
		\label{th:gaussian}
		Let $f$ be an arbitrary function generating $d$-dimensional outputs. Let $\epsilon \in(0,1)$ be arbitrary. For $c^2 > 2ln(1.25/\delta)$, the Gaussian Mechanism with parameter $\sigma \geq c\Delta_{2}f/\epsilon$ is $(\epsilon,\delta)$-differentially private.
	\end{lemma}
	
	\section{\xxx}
	\label{sec:jdp-vfl}
	This section presents our DP framework for VFL and analyzes its privacy and utility guarantees. Specifically, we present a new analysis of IR perturbation method for VFL-based GLM joint training. 
	Consider a VFL-based GLM joint training algorithm $\mathcal{A}$ with $T$ iterations. For iteration $t = 1,\ldots,T$, the $\mathbf{IR}$ is exchanged between single active party and passive parties to calculate loss and gradient.
	The joint training process won't stop until the model converges or it reaches the maximum iteration.
	
	We assume each passive party only exchanges $Sec[\mathbf{IR}]$ with active party, and active party exchanges $Sec[\mathbf{IR}]$ with all passive parties. In such an assumption, ``multi-passive-party'' setting can be deemed as a simple extension to ``single-active-passive-party'' setting. Algorithm \ref{alg:dp-vfl} takes ``single-active-passive-party'' setting as an example and gives our \xxx~algorithm. The $\mathbf{IR}_t^i$ denotes the intermediate result of the GLM in the $t$-th iteration of $i$-th party. 
	
	\begin{algorithm}[htb]
		\caption{\xxx}
		\label{alg:dp-vfl}{
			\begin{algorithmic}[1]
				\REQUIRE {Datasets $(\X^A, \y),{\X^B}$.
					Privacy loss $\epsilon,\delta \geq 0$.
					Number of epochs $e$. Number of mini-batches $r$. Norm clipping parameter $k>0$. Number of iterations $T=e*r$. Learning rate $\eta$. {Loss function $\ell(\cdot,\cdot)$ with Lipschitz constant $L$ and smooth parameters $\beta_{\theta},\beta_y$. Regularization parameter $\lambda$.} Target bound $k_y$.}
				\ENSURE {$\widehat{\w}_A$,$\widehat{\w}_B$}
				\STATE {Conduct entity resolution between parties to obtain common entities and then $r$.}
				\STATE Normalize data samples such that for all $i\in\{1,\ldots,n\}$, $\|\x_i\|_2\leq 1$.
				\STATE Initialize the iteration index $t=1$.
				\FOR{$u=1:e$}
				\FOR{$j=1:r$}
				\STATE Sample $t$-th mini-batch $\X_t$ with the sample indices $\s_t\subset\{1,\ldots,n\}$.
				\STATE $\mathbf{IR}_t^B = \X_t^B\w_t^{B}$, 
				\\$Sec[\mathbf{IR}_t^B] = \mathbf{IR}_t^B + \Z^B$,
				where $\Z^B\sim \mathcal{N}(\mathbf{0},\sigma_A^2\mathbf{I})$ is a sample of Gaussian distribution, and $\sigma_A = \sqrt{2\log(1.25/\delta)}(\Delta_2 ([\mathbf{IR}_t^B]_{t=1}^T) / \epsilon)$, where $\Delta_2 ([\mathbf{IR}_t^B]_{t=1}^T)$ is defined in Lemma~\ref{lemma:u-l2}.
				\\Passive party sends $Sec[\mathbf{IR}_t^B]$ to active party.  \COMMENT{ Passive Party}
				\STATE $\mathbf{IR}_t^A= [\frac{\partial \ell}{\partial \theta_{i,t}}|_{\theta_{i,t} =\x_i\w_t^A+Sec[\mathbf{IR}_t^B]_i}]_{i\in\s_t}$,
				where $\theta_{i,t}$ is defined in Eq.~\eqref{equ:obj-function}.
				\\$Sec[\mathbf{IR}_t^A] = \mathbf{IR}_t^A + \Z^A$, where $\Z^A\sim \mathcal{N}(\mathbf{0},\sigma
				_B^2\mathbf{I})$ is a sample of Gaussian distribution, and $\sigma_B = \sqrt{2\log(1.25/\delta)}(\Delta_2 ([\mathbf{IR}_t^A]_{t=1}^T) / \epsilon)$, where $\Delta_2 ([\mathbf{IR}_t^A]_{t=1}^T)$ is defined in Lemma~\ref{lemma:v-l2}.
				\\Active party sends $Sec[\mathbf{IR}_t^A]$ to passive party. \COMMENT{ Active Party}
				\STATE Compute gradient $\g_t^{A} = (\mathbf{IR}_t^A)\trans\X_j^{A}/b$. \COMMENT{Active Party} 
				\STATE Compute gradient $\g_t^{B} = (Sec[\mathbf{IR}_t^A])\trans\X_j^{B}/b$. \COMMENT{Passive Party}
				\STATE Update $\w_t^A = \mbox{Pen}(\w_t^A, \g_t^A,\eta,\lambda) $, $\w_t^B = \mbox{Pen}(\w_t^B, \g_t^B,\eta,\lambda)$.
				\STATE Norm clipping: $\w_t^A = \w_t^A/\max(1,\frac{\|\w_t^A\|_{2}}{k}),\w_t^B = \w_t^B/\max(1,\frac{\|\w_t^B\|_{2}}{k})$.
				\STATE Let $t = t+1$.
				\ENDFOR
				\ENDFOR
		\end{algorithmic}}
	\end{algorithm}
	\subsection{Algorithms}
	
	As shown in Algorithm \ref{alg:dp-vfl}, we introduce a differentially private method to calculate $Sec[\mathbf{IR}]$ to protect the training datasets. Unlike existing HE-based $Sec[\mathbf{IR}]$ calculation, \xxx~doesn't need to conduct polynomial approximation on loss function before HE can be applied. Instead, we can simply calculate $\mathbf{IR}$'s $\ell_2$ sensitivity and add Gaussian noise correspondingly. In the following sections, we will instantiate \xxx~framework by logistic regression and mathematically prove its multi-level privacy and utility guarantees. We will then evaluate our method in Section \ref{sec:evaluate}.
	
	\subsection{{Examples} of \xxx~Framework}
	\label{sec:instance}
	We take a popular machine learning method, $\ell_2$-regularized logistic regression with the $\ell_2$  regularization parameter $\lambda$, as an example of our \xxx~framework.  
	The objective function is:
	
	\begin{align}
	\label{equ:loss_function}\small
	\calL(\w) = \frac{1}{n}\sum_{i=1}^n\log(1 + \exp(-y_i\x_i\w)) + \frac{\lambda}{2}\|\w\|_2^{2}, \ y_i\in\{-1,+1\},\forall i.
	\end{align}
	
	Correspondingly, for $i\in\s_t$, each $i$-th entry of $\mathbf{IR}_t^A$ in Algorithm~\ref{alg:dp-vfl} equals
	\begin{align}
	\label{equ:vfl_gradient_agg}\small
	\frac{\partial \ell}{\partial \theta_{i,t}}\biggl|_{\theta_{i,t} =\x_i\w_t^A+Sec[\mathbf{IR}_t^B]_i} = \biggl(\frac{1}{1+\exp[-y_{i}(\x_i\w_t^A+Sec[\mathbf{IR}_t^B]_i)]}-1\biggr)y_{i}. 
	\end{align}
	
	Then the update operation with penalty in Algorithm~\ref{alg:dp-vfl} is $\mbox{Pen}(\w_t^{\cdot}, \g_t^{\cdot},\eta,\lambda)=\w_t^{\cdot}-\eta(\g_t^{\cdot}+\lambda\w_t^{\cdot})$.
	
	Other parameters are: $L=1,\beta_{\theta}=0.25,\beta_y=1.1,k_y=1$, which are defined in Section~\ref{sec:theory}.
	
	Examples for other loss functions of GLM and other types of penalties are deferred to Appendix~\ref{sec:extension}.
	
	
	\section{Theoretical Analyses}\label{sec:theory}
	
	This section provides privacy guarantees and utility analyses for Algorithm~\ref{alg:dp-vfl}. We first define notations and make some assumptions.
	
	\begin{definition}[$\Delta(\cdot)$]
		We define $\Delta v\coloneqq \|v-v'\|_2$, where $v$ and $v'$ are vectors from two adjacent datasets $D$ and $D'$, respectively, that differ by a single data instance. The changed data instance could be either a pair of $(\x_i^A,y_i^A)$ from the active party or a $\x_i^B$ from the passive party, $i\in\{1,\ldots,n\}$.
	\end{definition}
	
	
	\textbf{\emph{Variable spaces.}} We assume the spaces for model weights and data samples are bounded such that $\|\w\|_2\leq k$ and $\|\x_i\|_2\leq 1, \forall i$, which is natural from the normalization and norm clipping steps of Algorithm~\ref{alg:dp-vfl}. For each $i$, we assume that $y_i$ has a sub-exponential distribution with parameters $(\sigma,\nu)$ such that $|y_i|\leq k_y$ with high probability of at least
	\begin{align*}\small
	\mathbb{P}(|y_i|\leq k_y)\geq
	\left\{
	\begin{array}{ll}
	1-\exp(-{k_y^2}/{\sigma^2}), & 0\leq k_y\leq  {\sigma^2}/{\nu}\\
	1-\exp(-{k_y }/{\nu }), & k_y> {\sigma^2}/{\nu},
	\end{array}
	\right.
	\end{align*}
	which can cover a wide range of distributions, including the commonly-encountered Bernoulli, Poisson, and Gaussian distributions for logistic, Poisson, and least square regressions, respectively.

	
	
	\textbf{\emph{Properties of objective functions.}} 
	We assume that the loss function $\ell(\cdot,\cdot)$ in Eq.~\eqref{equ:obj-function} is $\gamma$-strongly convex, $\beta$-smooth, and $L$-Lipschitz-continuous (defined in Appendix~\ref{sec:loss_prop}) w.r.t. the model weights $\w$ and $\beta_{\theta}$-smooth w.r.t. the natural parameter $\theta_i,\forall i$. We also assume $\partial \ell /\partial \theta_i$ is $\beta_y$-Lipschitz-continuous w.r.t. $y_i,\forall i$. These properties can cover a wide range of loss functions, including logistic, least square, Huber, $\ell_2$ support-vector-machines loss, losses for Poisson and Gamma regression, etc.
	
	We first show that the differences resulted from adjacent datasets on $\w_t$s are bounded.
	
	\begin{lemma}[$(\Delta \w_t)^2$ recursion]
		\label{lemma:delta_w}
		Assume $\Delta\w_0=0$, then we have for any $\eta \leq \frac{2}{\beta + \gamma}$:
		\begin{align*}
		(\Delta\w_{t+1})^{2} \leq
		\begin{cases}
		(1 - \frac{2\eta(b-1)\beta\gamma}{b(\beta+\gamma)})(\Delta\w_{t})^{2} + 
		\frac{4\eta L}{b}\Delta\w_{t} + \frac{4\eta^{2}L^{2}}{b^{2}}, &\quad \text{if } t = j*b, j = 0,\cdots,e-1;
		\\(1-\frac{2\eta\beta\gamma}{\beta+\gamma})(\Delta\w_{t})^2,     & \quad \text{otherwise}.
		\end{cases}
		\end{align*}
	\end{lemma}
	
	
	\subsection{Privacy Guarantees}
	This section proves that the $Sec[\mathbf{IR}_t^A]$s and $Sec[\mathbf{IR}_t^B]$s in \xxx~algorithm prevent indirect information leakage from active party's raw data $(\X^{A},\y^{A})$ and passive party's raw data $\X^{B}$ respectively. Specifically we calculate the $\ell_2$-sensitivity of $[\mathbf{IR}_t^A]_{t=1}^T$ and $[\mathbf{IR}_t^B]_{t=1}^T$ and prove that the perturbations make our \xxx~algorithm joint differentially private.
	
	
	\begin{lemma}[$\ell_2$-sensitivity of $\mathbf{IR}_t^B$s]
		\label{lemma:u-l2}
		Let $T = e*r$ be the number of iterations, the $\ell_2$-sensitivity of $\mathbf{IR}_t^B$s in Algorithm \ref{alg:dp-vfl} is $\Delta_2 ([\mathbf{IR}_t^B]_{t=1}^T)=\sqrt{\frac{4L^2e^2T\eta^2}{b} + \frac{8k Le^2 \eta}{b} + 4k^2e}$.
	\end{lemma}
	
	The proofs of both Lemma \ref{lemma:delta_w} and \ref{lemma:u-l2} are deferred to Appendix \ref{proof:lm1} and Appendix {\ref{proof:lm2}}.
	
	
	
	\begin{lemma}[$\ell_2$-sensitivity of $\mathbf{IR}_t^A$s] 
		\label{lemma:v-l2}
		Let $T = e*r$ be the number of iterations, the $\ell_2$-sensitivity of $\mathbf{IR}_t^A$s in Algorithm \ref{alg:dp-vfl} is $\Delta_2 ([\mathbf{IR}_t^A]_{t=1}^T)=\sqrt{\frac{4\beta_{\theta}^2L^2e^2T\eta^2 }{b} +   \frac{8(\beta_{\theta}k+\beta_yk_y)\beta_{\theta}Le^2\eta }{b} + 4(\beta_{\theta}k+\beta_yk_y)^2e}$.
	\end{lemma}
	
	The proof of Lemma \ref{lemma:v-l2} is deferred to Appendix \ref{proof:lm3}.
	
	

	\begin{theorem}[DP]
		\label{th:dp-satisfy}
		Algorithm \ref{alg:dp-vfl} is $(\epsilon,\delta)$- differentially private w.r.t $[Sec[\mathbf{IR}_t^A]]_{t=1}^T$ and $[Sec[\mathbf{IR}_t^B]]_{t=1}^T$.
	\end{theorem}
	\begin{theorem}[JDP]
		\label{th:jdp-satisfy}
		Algorithm \ref{alg:dp-vfl} is $(\epsilon,\delta)$-joint differentially private w.r.t $[\w_t^A]_{t=1}^T$ and $[\w_t^B]_{t=1}^T$.
	\end{theorem}
	The proofs of Theorems~\ref{th:dp-satisfy} and~\ref{th:jdp-satisfy} are deferred to Appendices \ref{proof:dp-satisfy} and \ref{proof:jdp-satisfy}, respectively.
	
	Theorem~\ref{th:dp-satisfy} shows that through Algorithm \ref{alg:dp-vfl}, first, the perturbations in  $Sec[\mathbf{IR}_t^B]$s in passive party prevent active party from getting private information about raw data $\x_{i}^B$, by observing the changes in the sequence of $Sec[\mathbf{IR}_t^B]$s; then, the perturbations in  $Sec[\mathbf{IR}_t^A]$s in active party prevent passive party from getting private information about raw data $\x_{i}^A$ and $y_i^A$, by observing the changes in sequence of $Sec[\mathbf{IR}_t^A]$s. 
	On the other hand, Theorem~\ref{th:jdp-satisfy} further shows that through Algorithm \ref{alg:dp-vfl}, first, the perturbations in  $Sec[\mathbf{IR}_t^B]$s in passive party prevent active party from getting private information about raw data $\x_{i}^B$, by observing the changes in the sequence of $\w_t^A$s; then, the perturbations in  $Sec[\mathbf{IR}_t^A]$s in active party prevent passive party from getting private information about raw data $\x_{i}^A$ and $y_i^A$, by observing the changes in sequence of $\w_t^B$s.
	
	

	\subsection{Utility Analyses}
	We build utility analyses  for Algorithm~\ref{alg:dp-vfl}. Our utility analyses are built upon the error bounds of inexact proximal-gradient descent presented by \citet{schmidt2011convergence}. 

	Let $\w^* = \mathop{\arg\min}_{\w} \calL(\w)$, and $g(\cdot) = \|\cdot\|_2$. Without loss of generality, we assume that $\|\w_{0}-\w^{*}\|_2=O(k)$. Now, we present guarantees regarding both utility and runtime. 
	
	\begin{lemma}
		\label{lemma:error gradient}
		For all $t \in \{1,\ldots,T\}$, denote the gradient error caused by noise by ${\e}^{t} = \frac{1}{b}\sum_{i\in\s_t}\nabla \ell(\x_i\w_t,y_i)- [\mathbf{g}_t^A,\mathbf{g}_t^B]$, where $\mathbf{g}_t^A$ and $\mathbf{g}_t^B$ are defined in Algorithm~\ref{alg:dp-vfl}. It holds that $\|\e^{t}\| = O\left(\frac{\sqrt{\log(1.25/\delta)}}{\epsilon}\sqrt{\frac{\beta_{\theta}^2L^2e^2T\eta^2 }{b} +   \frac{2(\beta_{\theta}k+\beta_yk_y)\beta_{\theta}Le^2\eta }{b} + (\beta_{\theta}k+\beta_yk_y)^2e}\right)$.
	\end{lemma}
	
	
	\begin{theorem}
		\label{th:utility}
		For $\mathcal{E} = \calL (\frac{1}{T}\sum_{t=1}^{T}\w_{t} ) - \calL(\w^{*})$, we have, with high probability,
		\begin{equation}\small
		\begin{split}
		\mathcal{E} = O\biggl(\biggl[k\sqrt{\frac{\beta}{T}}  + 2\sqrt{\frac{T}{\beta}} \frac{\sqrt{\log(1.25/\delta)}}{\epsilon}\sqrt{\frac{\beta_{\theta}^2L^2e^2T\eta^2 }{b} +   \frac{2(\beta_{\theta}k+\beta_yk_y)\beta_{\theta}Le^2\eta }{b} + (\beta_{\theta}k+\beta_yk_y)^2e} \biggr]^{2}\biggr).    
		\end{split}
		\end{equation}
	\end{theorem}
	
	The proof of lemma \ref{lemma:error gradient} and theorem \ref{th:utility} are deferred to Appendix \ref{proof:lm4} and Appendix \ref{proof:th3}.
	
	\section{Experiments}
	\label{sec:evaluate}
	This section evaluates the proposed \xxx~method instantiated by a VFL-based regularized logistic regression task. We address three questions: (\textbf{Q1}) How is \xxx's privacy-accuracy tradeoff? (\textbf{Q2}) How does \xxx's hyper-parameters affect \xxx's accuracy under certain privacy requirement? (\textbf{Q3}) How is \xxx's runtime overhead?
	
	\subsection{Methods for Comparison}
	For the regularized logistic regression task, we evaluate five types of methods: \textbf{1)} single-party method, which trains a logistic regression model by active party and its dataset alone; \textbf{2)} traditional centralized non-FL method, which trains a regularized logistic regression model with all datasets located at a single party; \textbf{3)} idealized non-private VFL method, which jointly trains a logistic regression model, with datasets partitioned at two parties, by exchanging intermediate result $\mathbf{IR}$ directly; \textbf{4)} HE-VFL method, which jointly trains a logistic regression model, with datasets partitioned at two parties, by a) approximating loss and gradient to low-degree polynomial representations, and b) exchanging HE-based polynomial $Sec[\mathbf{IR}]$ between parties; \textbf{5)} our \xxx~method, which is similar to idealized non-private VFL method except that differentially private $Sec[\mathbf{IR}]$ is exchanged between parties.
	
	We implement three VFL-based methods in FATE-1.3 \cite{fate}, an open source platform for VFL research. For single-party and centralized non-FL methods, we leverage the logistic regression classifier from sklearn. We use three real-world datasets from UCI Machine Learning Repository\cite{Dua:2019} for our evaluation, detailed in Table \ref{tab:datasets}.
	We split the datasets vertically into two sub-datasets with comparable amount of attributes and distribute them to active party and passive party respectively. 
	We use test accuracy as our evaluation metric. All experimental data is average of 10 runs.
	
	\begin{table}[htbp]
		\centering
		\caption{Datasets for Active Party and Passive Party}
		\label{tab:datasets}
		{\small
			\begin{tabular}{l|c|c|c|c}
				\hline \textbf{Datasets}& \textbf{Task}& \textbf{\# of Samples}& \textbf{\# of Attributes (Active)}& \textbf{\# of Attributes (Passive)} \\
				\hline Breast\cite{breast} & Binary Classification& 569& 11 & 20\\
				\hline Credit\cite{credit}& Binary Classification& 30000& 14 & 10\\
				\hline Adult\cite{adult}& Binary Classification& 32561& 7& 8\\			
				\hline
		\end{tabular}}
	\end{table}
	
	\subsection{Implementation Details}
	\label{sec:exp_setting}
	We set $\lambda = 0.001$ as default for all our datasets.
	The epoch number $e$ and weight constraint $k$ are \xxx's two important hyper-parameters which will affect  $\mathbf{IR}$'s $\ell_2$ sensitivity. Normally the larger the sensitivity value, the larger the noise needed to maintain differentially private, and the lower the accuracy. We tune these hyper-parameters for the best privacy-accuracy tradeoffs. Specifically, we tune $e$ in $[5, 15]$ and $k$ in $[0.1,1]$ using 5-fold cross-validation method on the training datasets.
	We set $\delta = 0.01$  according to the work of \citet{boyd2015differential}.
	
	
	
	
	
	\subsection{Privacy-Accuracy Tradeoff}
\begin{figure}[ht]
		\begin{subfigure}{.33\textwidth}
			\centering
			\includegraphics[width=\linewidth]{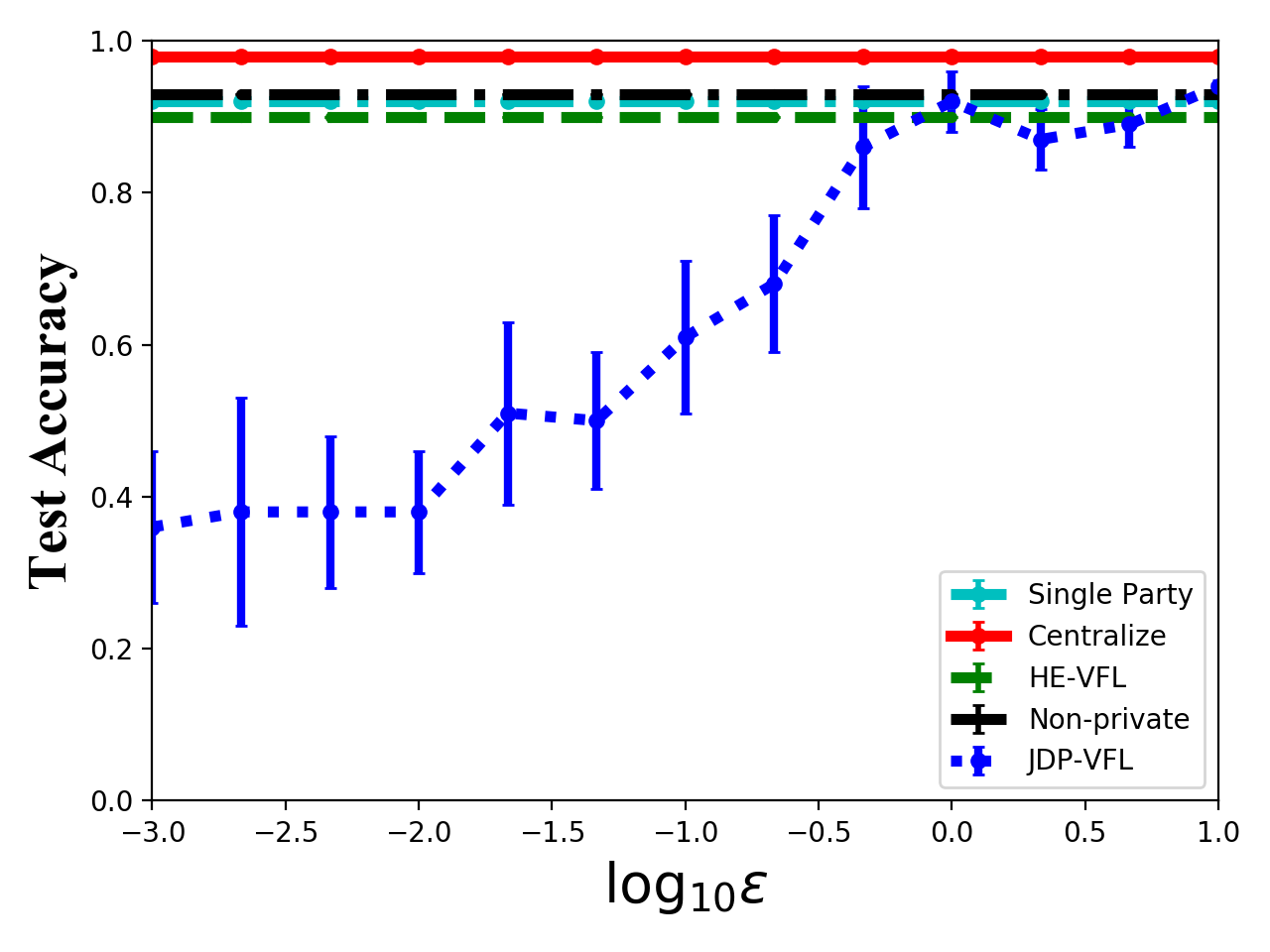}
			\caption{Breast}
		\end{subfigure}
		\begin{subfigure}{.33\textwidth}
			\centering
			\includegraphics[width=\linewidth]{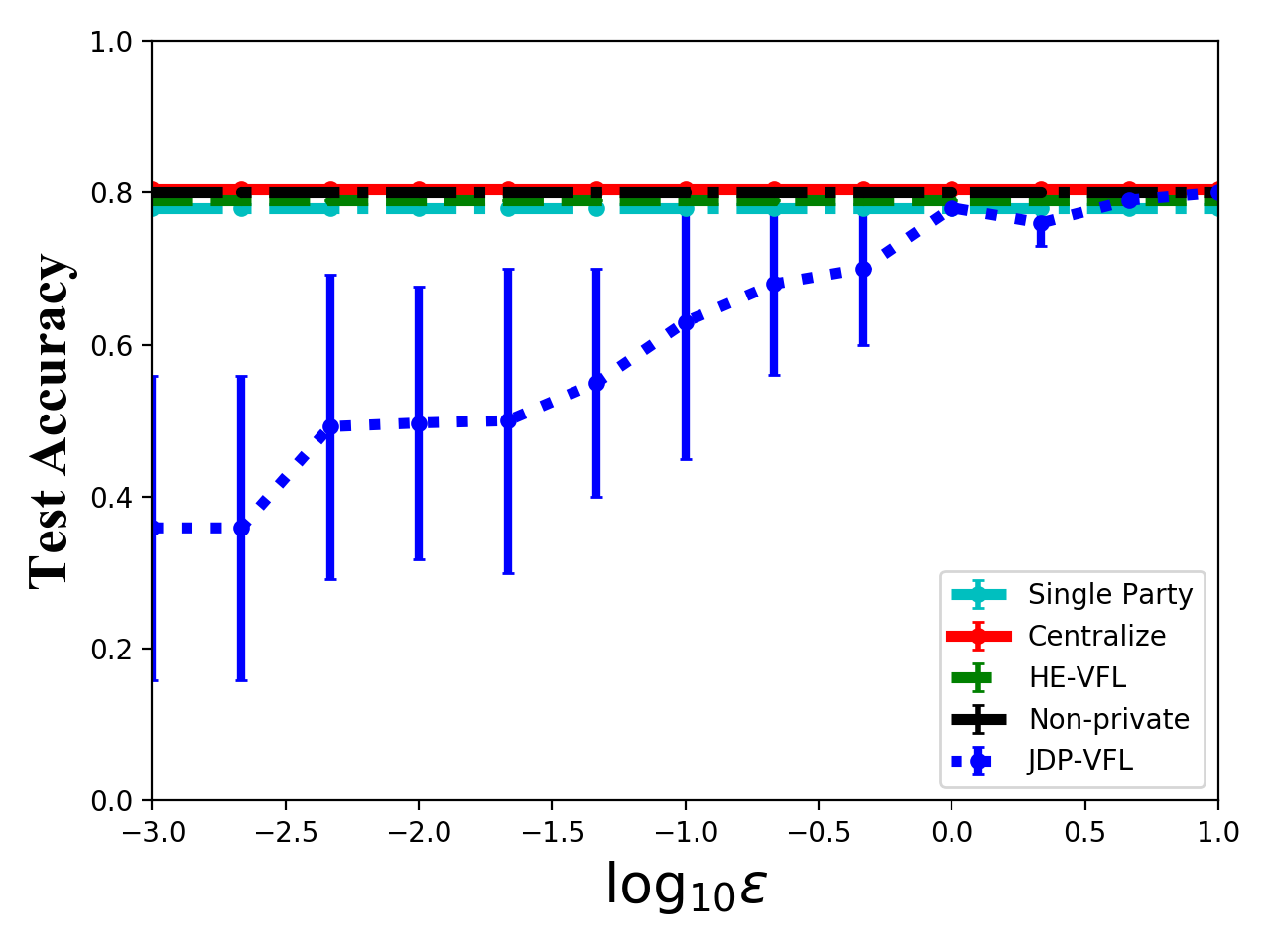}
			\caption{Credit}
		\end{subfigure}
		\begin{subfigure}{.33\textwidth}
			\centering
			\includegraphics[width=\linewidth]{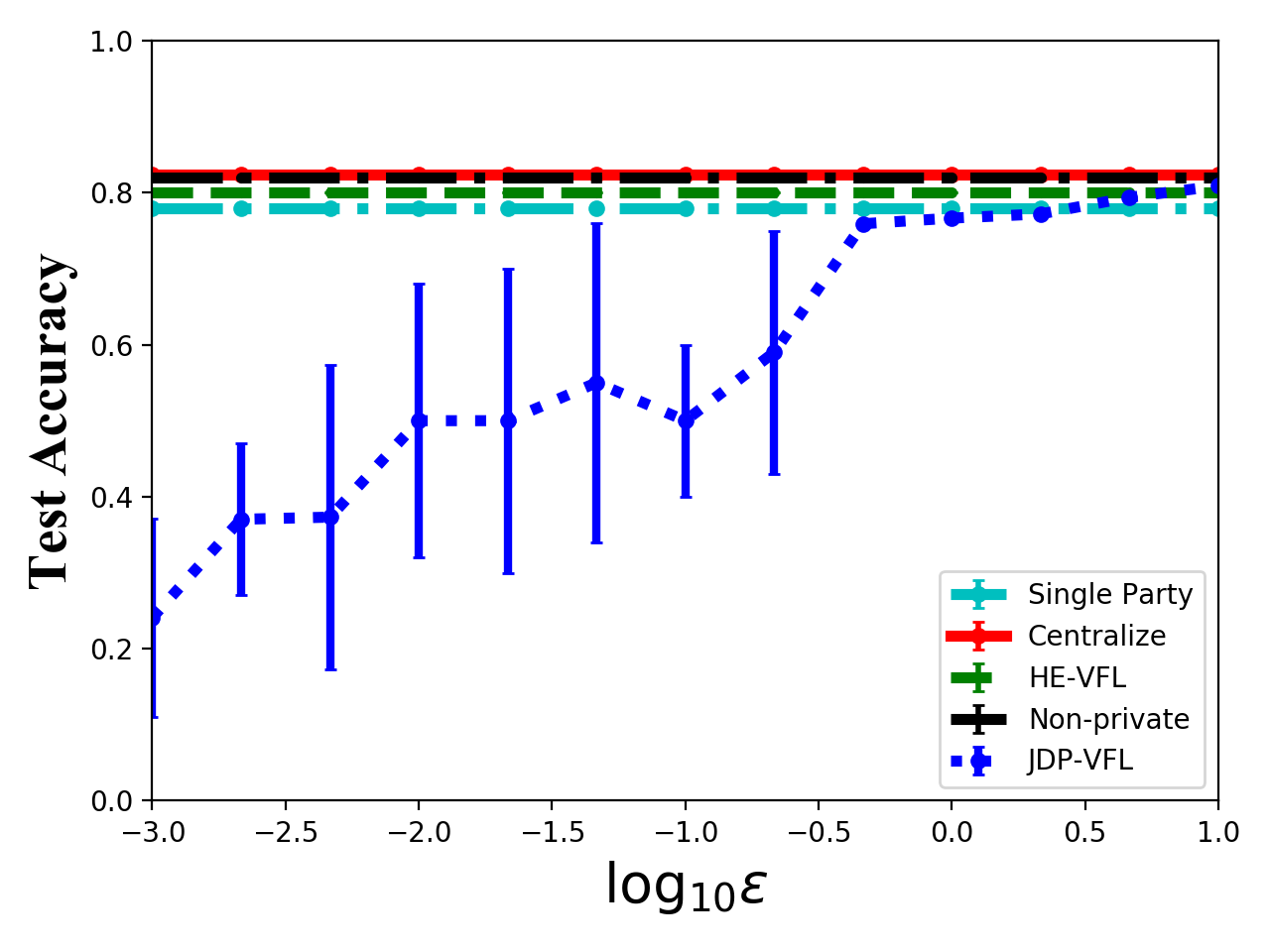}
			\caption{Adult}
		\end{subfigure}
		\caption{\xxx's privacy-accuracy tradeoff results using public datasets. We set mini-batch size $b$ = 3200, $\lambda = 0.001$, epoch number $e$ = 10, and weight constraint $k$ = 1.}
		\label{exp:acc trade off}
	\end{figure}

	First we study \xxx's tradeoff between the privacy requirement in specific range and the accuracy of a binary classification task. By adjusting the parameters mentioned in Section \ref{sec:exp_setting}, Figure \ref{exp:acc trade off} reports the \xxx's results on privacy and accuracy tradeoff.
	
	From the results we can see that the best accuracy result \xxx~could achieve within the given privacy $\epsilon$ range in $[0.001, 10]$ is comparable to single-party method, centralized method, idealized non-private VFL method, and HE-VFL method which is deemed as lossless. This indicates that \xxx~could achieve high accuracy when privacy budget is sufficient, e.g. above $10$, but low accuracy, only half of the lossless accuracy, when privacy budget is very tight, e.g. below $0.1$. In practice using our \xxx~method, we set $\epsilon=1$ which achieves acceptable accuracy-privacy tradeoffs. The privacy-accuracy tradeoff evaluation result on the full range of $\epsilon$ are shown in the supplementary material(Appendix \ref{appendix:exp}).
	
	\subsection{Effects of Hyper-parameters}
	
\begin{figure}[ht]
		\centering
		\begin{subfigure}{.3\textwidth}
			\centering
			\includegraphics[width=\linewidth]{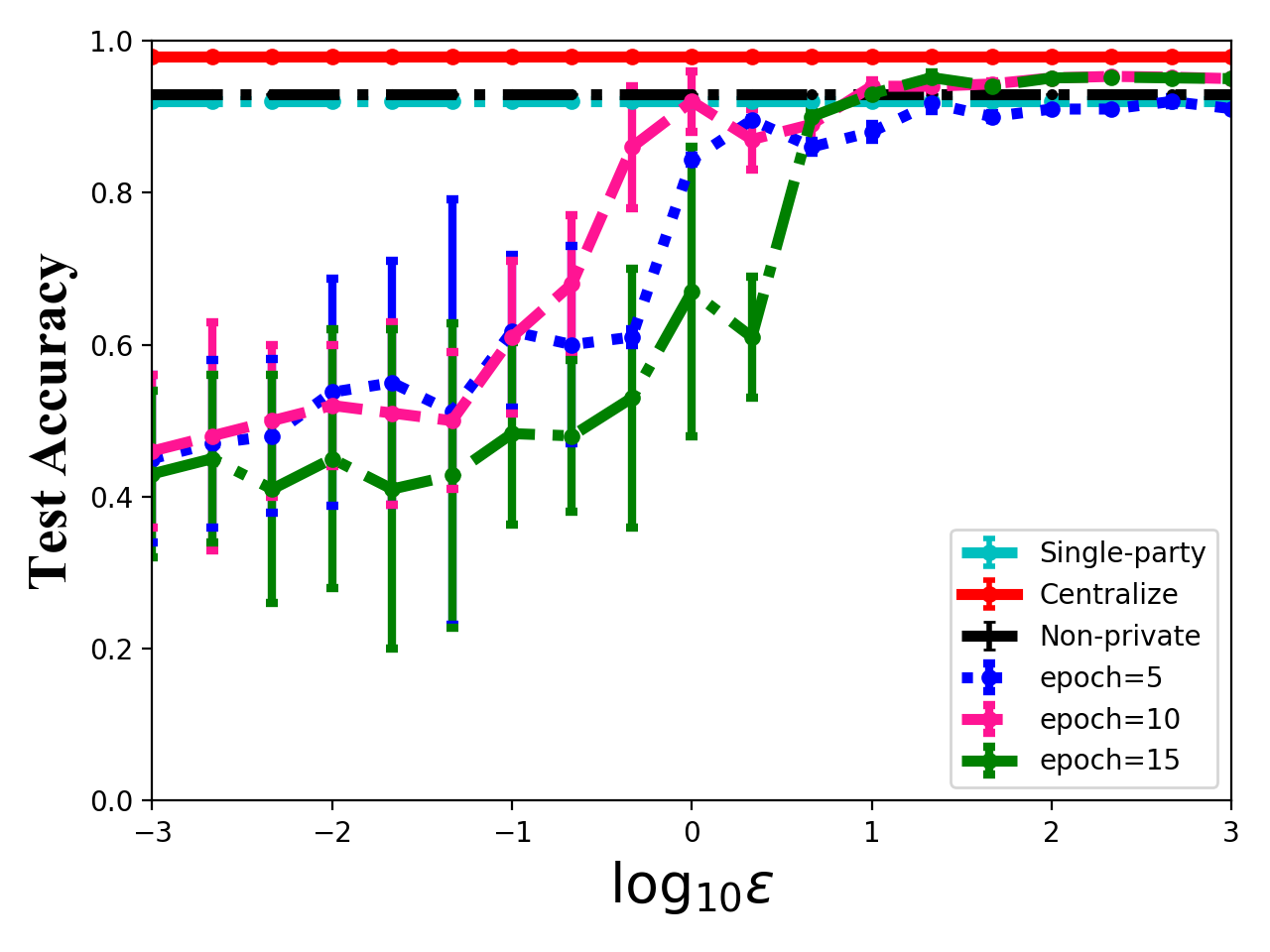}
			\caption{}
		\end{subfigure}
		\begin{subfigure}{.3\textwidth}
			\centering
			\includegraphics[width=\linewidth]{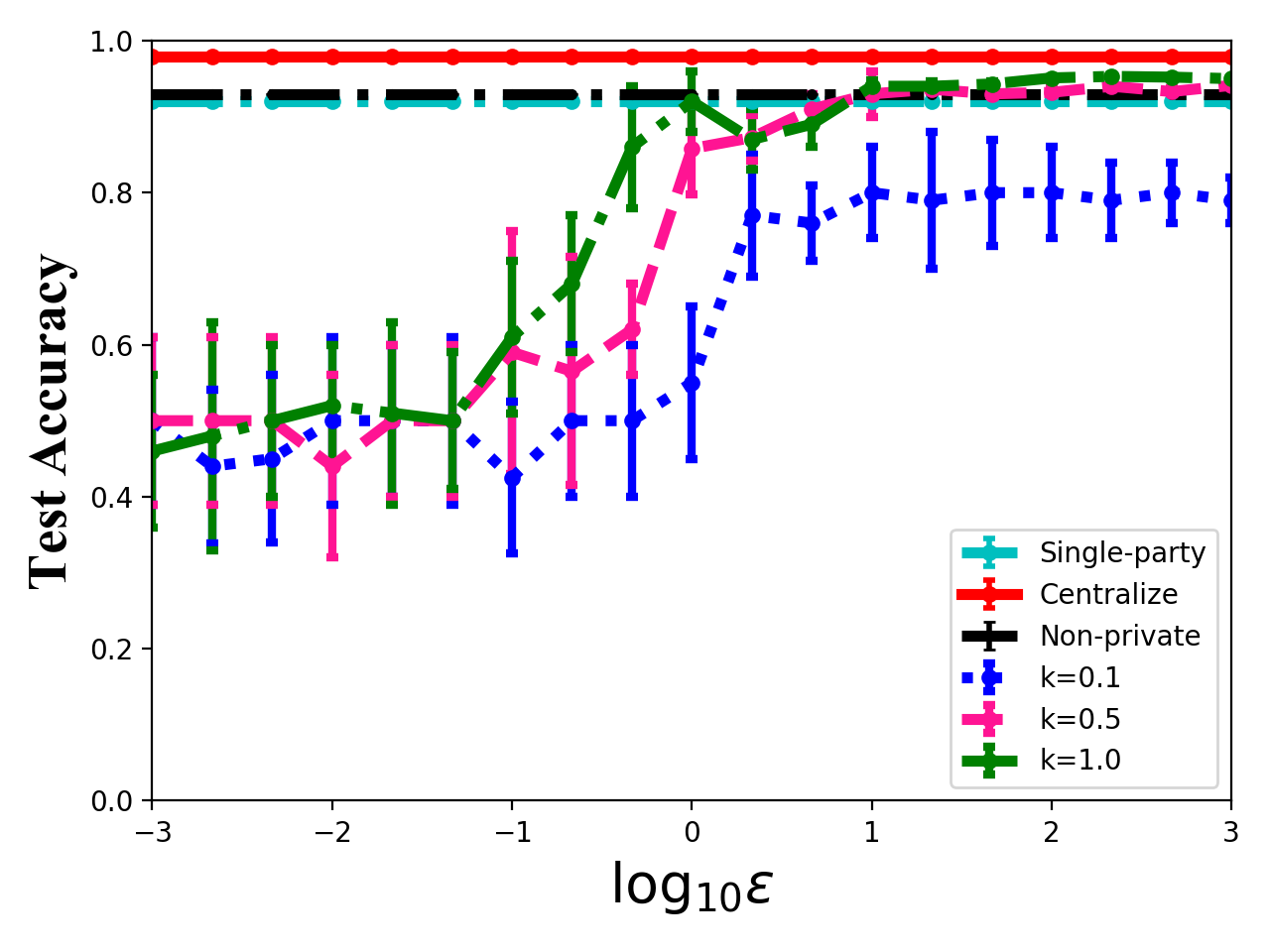}
			\caption{}
		\end{subfigure}
		\caption{Privacy-accuracy tradeoff when tuning hyper-parameters of \xxx~on Breast dataset.}
		\label{exp:param}
	\end{figure}

	Then we study how the hyper-parameters affect its accuracy under certain privacy requirement. For each hyper-parameter under a given range, e.g. range in Section \ref{sec:exp_setting}, we choose three values, e.g., lower bound, upper bound, and a middle value, to study the privacy-accuracy tradeoffs. Figure \ref{exp:param} shows the results of tuning hyper-parameters epoch number $e$ and weight constraint $k$. 
	
	From the results we can see that under different privacy budget, \xxx~could achieve different accuracy results with different hyper-parameter value. For example, by changing $e$ from $5$ to $15$ under $\epsilon=1$, \xxx's accuracy drops from $0.9$ to around $0.6$. Similarly, by changing $k$ from $0.1$ to $0.5$, the accuracy increases from $0.5$ to around $0.9$. The reason behind this is that \xxx's hyper-parameters affect $\mathbf{IR}$'s $\ell_2$ sensitivity, as analyzed in Lemma \ref{lemma:u-l2} and Lemma \ref{lemma:v-l2}. This also indicates that under certain privacy budget, the hyper-parameter tuning should be targeting at minimizing $\mathbf{IR}$'s $\ell_2$ sensitivity.
	
	\subsection{Runtime Overhead}
	\begin{figure}[ht]
		\centering
		\begin{subfigure}{.3\textwidth}
			\centering
			\includegraphics[width=\linewidth]{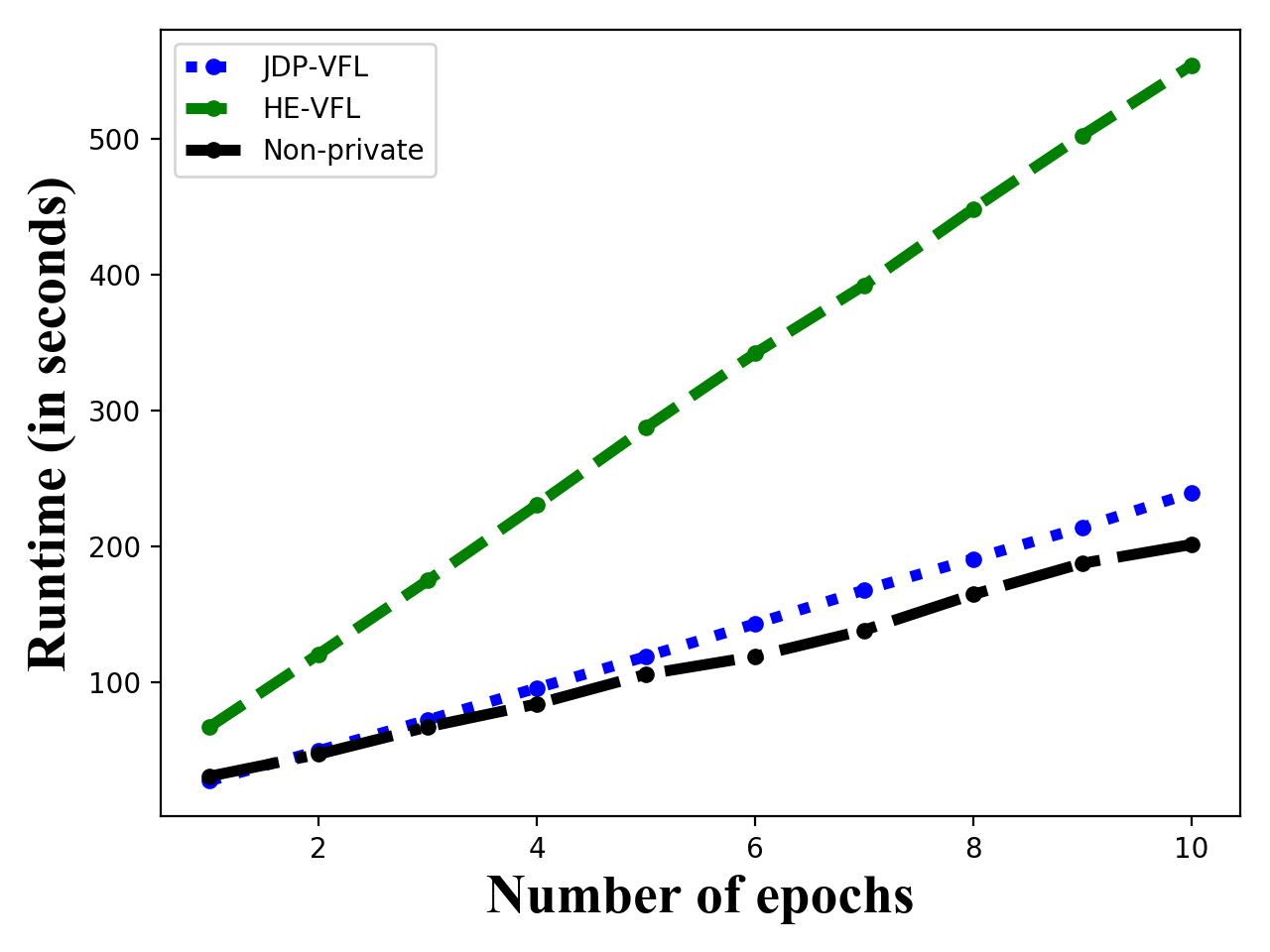}
			\caption{}
		\end{subfigure}
		\begin{subfigure}{.3\textwidth}
			\centering
			\includegraphics[width=\linewidth]{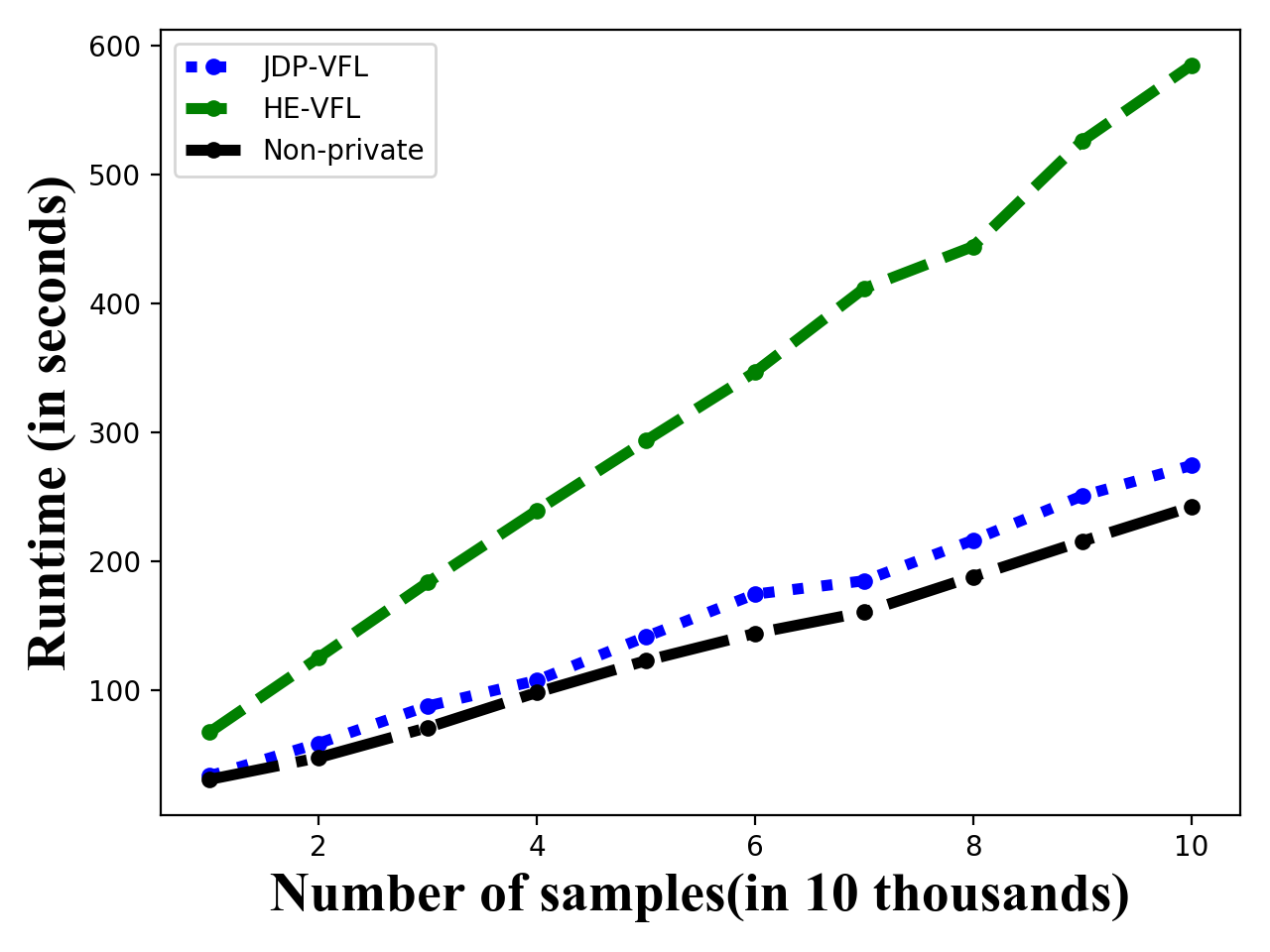}
			\caption{}
		\end{subfigure}
		\caption{The runtime overhead of idealized non-private VFL, HE-VFL, and \xxx. In (a), we change the number of epochs from 1 to 10 and set the size of dataset to 10000. In (b), we vary the dataset size from 10,000 to 100,000 and set the number of epochs to 1.}
		\label{exp:overhead}
	\end{figure}
	
	Finally we study \xxx's runtime overhead. We mainly compare runtime overhead of idealized non-private VFL, HE-VFL, and \xxx. From the result in Figure \ref{exp:overhead} we can see that \xxx~achieve the similar runtime overhead as the idealized non-private VFL, whereas HE-VFL has the largest runtime overhead, roughly $2 \sim 3$ times slower than both non-private VFL and JDP-VFL. More importantly, we can see the runtime overheads of three VFL methods are proportional to the number of epochs and samples. This result strongly indicates that \xxx~could significantly save joint training time under VFL setting where both parties have large amount of data samples.
	
	\section{Conclusions}
	Privacy-preserving vertical federated learning (VFL) is one of the effective solutions for enterprise-level data collaborations while respecting data privacy. 
	However, the commonly used HE-based VFL suffers from the increased cost in memory and processing time when the number of training samples is huge.
	This paper studies this issue and presents \xxx, the first differentially private framework for VFL.
	By analyzing the sensitivity of VFL's intermediate result (IR) and conducting perturbation of IR directly within each training iteration, \xxx~
	doesn't need the Taylor approximation step and the third-party collaborator of HE-VFL, thus \xxx~is easy to deploy. We mathematically prove that \xxx~provides multi-level privacy and utility guarantees. Experimental results show the effectiveness of \xxx. 
	
	\newpage
	\section*{Broader Impact}
	As any federated learning related research which trades communication efficiency and training time for data privacy and thus has an impact on energy consumption, our work, which focuses on acceleration of the vertical federated training process without compromising privacy guarantees, is no exception. Specifically, this work has a positive impact on society to respect data privacy, by complying with government regulations like GDPR\cite{10.5555/3152676}, when conducting collaborative machine learning tasks on personal data or enterprise data. At the same time, this work may have some negative consequences: 1) our work uses differentially private method and mathematical proofs to replace the time-consuming and memory-consuming homomorphic encryption based privacy-preserving federated training process, thus it may be difficult, when privacy budget is abnormally tight, to gain a lossless joint model as homomorphic encryption based vertical federated learning (VFL); 2) the low performance joint model, under abnormally tight privacy budget, may fail to deliver the expected outcomes for data collaboration between organizations; 3) our method inherits the same limitation of existing VFL, which requires datasets of organizations have to share the same or similar sample space but differ in feature space. Furthermore, we should be cautious of the fact that our method only protects enterprise data, and how enterprise data is collect from personal data is beyond the scope of this work. Finally, this work does leverage biases in the data, which is the primary task of this work.
	
	\bibliographystyle{abbrvnat}
	\bibliography{ref} 
	
	\newpage
	\appendix

	\section{Definitions of Properties for Loss Functions}\label{sec:loss_prop}
	\begin{definition}
		\label{df:loss}
		Let $f: \mathcal{W} \rightarrow \mathbb{R}$ be a function, where $\mathcal{W}$ is a hypothesis space equipped with the standard inner product and $\ell_2$ norm $\|\cdot\|:$
		
		1) $f$ is $L$-Lipschitz if for any $u, v \in \mathcal{W}$,
		\begin{align*} \|f(u) - f(v)\| \leq L\|u - v\|; \end{align*}
		2) $f$ is $\beta$-smooth if
		\begin{align*} \|\nabla f(u) - \nabla f(v)\| \leq \beta\|u-v\|; \end{align*}
		3) $f$ is $\gamma$-strongly convex if
		\begin{align*} f(u)\geq f(v) + \langle\nabla f(v), u - v\rangle + \frac{\gamma}{2}\|u - v\|^{2}. \end{align*}
	\end{definition}

	
	
	
	\textbf{Post-Processing immunity}. This property helps us safely use the output of a differentially private algorithm without additional information leaking, as long as we do not touch the dataset $D$ again.
	\begin{property}[Post-Processing immunity. Proposition 2.1 in~\citet{dwork2014algorithmic}]\label{property:post-processing}
		
		Let algorithm $\mathcal{A}_1(\mathcal{B}_1): D \rightarrow I_1 \in \mathcal{R}$ be an $(\epsilon,\delta)$ - differential privacy algorithm, and let $f: \mathcal{R} \rightarrow \mathcal{R}'$ be an arbitrary mapping. Then, algorithm $\mathcal{A}_2(\mathcal{B}_2):  D \rightarrow I_2 \in \mathcal{R}'$
		is still $( \epsilon, \delta )$ - differentially private, i.e., for any set $\calS\subseteq \mathcal{R}$,
		\begin{align*}
		\mathbb{P}(I_2 \in \mathcal{S} \mid \mathcal{B}_2 = D  )
		\leq  e^{\epsilon}\mathbb{P}(I_2 \in \mathcal{S} \mid \mathcal{B}_2 = D' ) + \delta.
		\end{align*}
	\end{property}
	
	\section{Proof of Results In The Main Text}
	
	\subsection{Proof of Lemma \ref{lemma:delta_w} [$(\Delta \w_t)^2$ recursion]}
	\label{proof:lm1}
	\begin{proof}
		Let $S$ denote the mini-batch of data with the sample indices $\s\in\{1,\ldots,n\}$ and $|\s|=b$, and let $\w_t$ denote the model weights in the $t$-th step of \xxx's joint training described in Algorithm \ref{alg:dp-vfl}. {Let $\mathcal{F}(\w_t,S)=\frac{1}{b}\sum_{i\in\s_t}\ell(\w_t,\x_i)$ denote the average loss function for $S$.} Let $S'$ be the ``neighboring data'' of $S$, and let $\w'_t$ be the model weights trained from $\mathcal{S}'$. To calculate the recursion of $(\Delta \w_t)^2$, consider two cases of $S$ and $S'$: 1) $S$ is not changed in the $t$-th step of \xxx, thus $S = S'$; 2) $S$ and $S'$ are neighboring data differing in just one element $\x_i \rightarrow \x'_i$ or $(\x_i,y_i) \rightarrow (\x'_i,y'_i)$. We omit $y'_i$ when $(\x_i,y_i)$ change to $(\x'_i,y'_i)$ for short.
		Following the proof of Lemma 3.7.3 of~\citet{hardt2015train} we have:
		
		\textbf{Case 1)}: no data instance in $S$ is changed, we have
		\begin{align*}
		(\Delta\w_{t+1})^2 = &\|\w_{t+1} - \w_{t+1}'\|^2
		\\= &\|\w_t - \eta\nabla \mathcal{F} (\w_t,S) - \w_t’ + \eta\nabla \mathcal{F} (\w_t',S)\|^{2}
		\\= &\|\w_t - \w_t’\|^2 + \eta^2\|\nabla \mathcal{F} (\w_t',S) - \nabla \mathcal{F} (\w_{t}, S)\|^{2} \\
		&- 2\eta \langle \w_{t} - \w_{t}', \nabla \mathcal{F}(\w_t',S)- \nabla \mathcal{F}(\w_{t}, S) \rangle
		\\= &\|\w_t - \w_t'\|^2 + \eta^{2}\|\frac{1}{b}\sum_{i=1}^{b}(\nabla \ell(\w_{t},\x_{i}) - \nabla \ell(\w_{t}',\x_{i}))\|^{2} \\
		&- 2\eta\langle \w_t - \w_t',\frac{1}{b}\sum_{i=1}^{b}(\nabla \ell(\w_{t},\x_{i}) - \nabla \ell(\w_{t}',\x_{i}))\rangle
		\\\leq& (1-2\frac{\eta\beta\gamma}{\beta+\gamma})(\Delta\w_{t})^2 - 
		(\frac{2\eta}{\beta+\gamma}-\eta^{2})\|\frac{1}{b}\sum_{i=1}^{b}(\nabla \ell(\w_{t},\x_{i}) - \nabla \ell(\w_{t}',\x_{i}))\|^{2}
		\\\leq&(1-2\frac{\eta\beta\gamma}{\beta+\gamma})(\Delta\w_{t})^2,
		\end{align*}
		where the first inequality, using the following inequality:
		\begin{align*}
		\langle \w_{t} - \w_{t}', \nabla \ell(\w_{t},\x) - \nabla \ell(\w_{t}’, \x) \rangle 
		\geq \frac{\beta\gamma}{\beta+\gamma}\|\w_{t} - \w_{t}’\|^{2} + \frac{1}{\beta+\gamma}\|\nabla \ell(\w_{t},\x) - \nabla \ell(\w_{t}', \x)\|^{2}.
		\end{align*}
		
		\textbf{Case 2)}: one data instance in $S$ is changed, we have
		\begin{align*}
		(\Delta\w_{t+1})^2 = &\|\w_{t+1} - \w_{t+1}’\|^2\\
		= &\|\w_t - \eta\nabla \mathcal{F} (\w_t,S) - \w_t' + \eta\nabla \mathcal{F} (\w_t’,S’)\|^{2}\\
		= &\|\w_t - \w_t'\|^2 + \eta^2\|\nabla \mathcal{F} (\w_t',S') - \nabla \mathcal{F} (\w_{t}, S)\|^{2} - 2\eta \langle \w_{t} - \w_{t}', \nabla \mathcal{F}(\w_t',S') \\
		&- \nabla \mathcal{F}(\w_{t}, S) \rangle\\
		\leq& \|\w_t - \w_t'\|^2 + \eta^{2}\|\frac{1}{b}\sum_{i=1}^{b-1}(\nabla \ell(\w_{t},\x_{i}) \\
		&- \nabla \ell(\w_{t}',\x_{i}))\|^{2} - 2\eta\langle \w_t - \w_t',\frac{1}{b}\sum_{i=1}^{b-1}(\nabla \ell(\w_{t},\x_{i}) - \nabla \ell(\w_{t}',\x_{i}))\rangle\\
		&+\eta^{2}\|\frac{1}{b}(\nabla \ell(\w_{t},\x_{i}) - \nabla \ell(\w_{t}',\x_{i}'))\|^{2} - 2\eta\langle \w_t - \w_t',\frac{1}{b}(\nabla \ell(\w_{t},\x_{i}) - \nabla \ell(\w_{t}',\x_{i}'))\rangle\\
		\leq& (1 - \frac{2\eta(b-1)\beta\gamma}{b(\beta+\gamma)})(\Delta\w_{t})^{2}-(\frac{2\eta}{\beta+\gamma}-\eta^{2})\|\frac{1}{b}\sum_{i=1}^{b-1}(\nabla \ell(\w_{t},\x_{i}) - \nabla \ell(\w_{t}',\x_{i}))\|^{2}\\
		&+\eta^{2}\|\frac{1}{b}(\nabla \ell(\w_{t},\x_{i}) - \nabla \ell(\w_{t}',\x_{i}'))\|^{2} - 2\eta\langle \w_t - \w_t',\frac{1}{b}(\nabla \ell(\w_{t},\x_{i}) - \nabla \ell(\w_{t}',\x_{i}'))\rangle\\
		\leq& (1 - \frac{2\eta(b-1)\beta\gamma}{b(\beta+\gamma)})(\Delta\w_{t})^{2} + 
		\frac{4\eta L}{b}\Delta\w_{t} + \frac{4\eta^{2}L^{2}}{b^{2}}.
		\end{align*}
		In summary, the recursion about $\Delta\w_t$ is:
		\begin{align*}
		(\Delta\w_{t+1})^{2} \leq
		\begin{cases}
		(1 - \frac{2\eta(b-1)\beta\gamma}{b(\beta+\gamma)})(\Delta\w_{t})^{2} + 
		\frac{4\eta L}{b}\Delta\w_{t} + \frac{4\eta^{2}L^{2}}{b^{2}}, &\quad \text{if } t = j*b, j = 0,\cdots,e-1;
		\\(1-\frac{2\eta\beta\gamma}{\beta+\gamma})(\Delta\w_{t})^2 ,    & \quad \text{otherwise}.
		\end{cases}
		\end{align*}
		
		From the above recursion we can know that,
		$(\Delta \w_{t+1})^{2} \leq (\Delta\w_{t})^{2}$ for \textbf{Case 1)} and $(\Delta \w_{t+1})^{2} \leq (\Delta \w_{t} + \frac{2\eta L}{b})^2$ in \textbf{Case 2)}.  Consider the assumption $\Delta \w_0 = 0$, then we have $(\Delta \w_{T})^2 \leq (\frac{2e\eta L}{b})^2$.
	\end{proof}
	\subsection{Proof of Lemma \ref{lemma:u-l2} [$\ell_2$-sensitivity of $\mathbf{IR}_t^B$s]}
	\label{proof:lm2}
	\begin{proof}
		Let $b$ be the mini-batch size, $r$ be the number of mini-batches, $e$ be the number of epochs, $T = r*e$, $\|\x\| \leq 1$, $\|\w^{B}\| \leq k$. $S$ and $S'$ are neighboring data differing in just one element $\x_i^{B} \rightarrow {\x'_i}^{B}$. Similar to \ref{proof:lm1}, consider two cases of $S$ and $S'$: 1) $S$ is not changed in the $t$-th step of \xxx, thus $S = S'$; 2) $S$ and $S'$ are neighboring data differing in just one element $\x_i^{B} \rightarrow {\x'_i}^{B}$.
		\\First, consider the $\Delta_2 (\mathbf{IR}_t^B)$ in passive party's single step:
		\\\textbf{Case 1)}: $\x_i^{B}$ is unchanged:
		\begin{align*}
		\Delta(\mathbf{IR}_t^B) &= \sqrt{\sum_{i=1}^{b}(\w_t^{B}\x_{i}^{B} - {\w'_t}^{B} \x_{i}^{B})^2} 
		\\&\leq \sqrt{\sum_{i=1}^{b}(\|\x_{i}^{B}\|\|\w_t^{B}-\w_t^{B'}\|)^2}
		\\&\leq \sqrt{b(\Delta\w_t)^{2}}.
		\end{align*}
		\textbf{Case 2)}: $\x_i^{B} \rightarrow {\x'_i}^{B}$:
		\begin{align*}
		\Delta(\mathbf{IR}_t^B) &= \sqrt{\sum_{i'\neq i}(\w_t^{B}\x_{i'}^B - {\w'_t}^{B}\x_{i'}^B)^2+(\w_t^{B}\x_{i} - {\w'_t}^{B}{\x'_{i}}^B)^2}
		\\&\leq\sqrt{\sum_{i'\neq i}(\w_t^{B}\x_{i'}^B - {\w'_t}^{B}\x_{i'}^B)^2+(|\w_t^{B}\x_{i} - {\w_t^{B}}{\x'_{i}}^B|+|\w_t^{B}\x_{i} - {\w'_t}^{B}\x_{i}|)^2}
		\\&\leq \sqrt{(b-1)(\Delta\w_t)^{2}+(2k+\Delta\w_t)^2}
		\\&=\sqrt{b(\Delta\w_t)^{2} + 4k\Delta\w_t + 4k^2}.
		\end{align*}
		Then, consider multiple steps for the passive party:
		\begin{align*}
		\Delta([\mathbf{IR}_t^B]_{t=1}^T) &= \sqrt{\sum_{t=1}^{T}(\Delta(\mathbf{IR}_t^B))^2}
		\\&\leq \sqrt{(T-e)b(\Delta\w_T)^{2}+e(b(\Delta\w_T)^{2} + 4k\Delta\w_T + 4k^2)}.
		\end{align*}
		Combining the proof of lemma \ref{lemma:delta_w}, we have:
		\begin{align*}
		\Delta([\mathbf{IR}_t^B]_{t=1}^T) 
		&\leq \sqrt{\frac{T(2e\eta L)^2}{b} + \frac{8ke^2 \eta L}{b} + 4ek^2}\\
		&= \sqrt{\frac{4L^2e^2T\eta^2}{b} + \frac{8k Le^2 \eta}{b} + 4k^2e}\\
		&=\Delta_2([\mathbf{IR}_t^B]_{t=1}^T).
		\end{align*}
		
	\end{proof}
	\subsection{Proof of Lemma \ref{lemma:v-l2} [$\ell_2$-sensitivity of $\mathbf{IR}_t^A$s]}
	\label{proof:lm3}
	\begin{proof}
		Consider $\mathbf{IR}_t^A = [h(\w_t,\x_i,y_i)]_{i\in \s_t}$, where $\s_t$ is the indices of the mini-batch for the $t$-th step, and
		\begin{align*}
		h(\w_t,\x_i,y_i) = \frac{\partial \ell}{\partial \theta_{i,t}}\biggl|_{\theta_{i,t} =\x_i\w_t^A+Sec[\mathbf{IR}_t^B]_i}. 
		\end{align*}
		Let $h(\cdot)$ denote $h$ w.r.t. the enclosed variable.
		
		Assume there exist constants $\beta_y,\beta_{\x},\beta_{\w}>0$, such that for all $y,y',\x,\x',\w,\w'$
		\begin{align*}
		&|h(y)-h(y')|\leq \beta_y|y-y'|\\
		&\|h(\x)-h(\x')\|\leq \beta_{\x}\|\x-\x'\|\\
		&\|h(\w)-h(\w')\|\leq \beta_{\w}\|\w-\w'\|.
		\end{align*}
		For generalized linear model, $\theta=\x\w$. Because $\ell(\cdot,\cdot)$ is $\beta_{\theta}$-smooth w.r.t. $\theta$, then 
		\begin{align*}
		&\frac{\partial \ell}{\partial \theta \partial \x}
		=\frac{\partial \ell}{\partial \theta \partial \theta}\frac{\partial \theta}{ \partial \x} \leq \beta_{\theta}\|\w\|\leq \beta_{\theta}k, \\
		&\frac{\partial \ell}{\partial \theta \partial \w}
		=\frac{\partial \ell}{\partial \theta \partial \theta}\frac{\partial \theta}{ \partial \w} \leq \beta_{\theta}\|\x\|\leq \beta_{\theta}. 
		\end{align*}
		Therefore, we have $\beta_{\x}=\beta_{\theta}k,\beta_{\w}=\beta_{\theta}$.
		
		First, consider one step for the active party. We consider two cases:
		
		\textbf{Case 1)}: no instance is changed, we have:
		\begin{align*}
		\Delta([h(\w_t,\x_i,y_i)]_{i\in \s_t}) &= \sqrt{\sum_{i\in\s_t}(h(\w_t,y_i)-h(\w_t',y_i))^2}
		\\&\leq \sqrt{b(\beta_{\w}\Delta\w_t)^2}.
		\end{align*}

		
		\textbf{Case 2)}: the $i$-th instance is changed, i.e., $(\x_i,y_i) \rightarrow (\x'_i,y'_i)$ we have:
		\begin{align*}
		&[\Delta([h(\x_i,\w_t,y_{i})]_{i \in \s_t})]^2 \\
		= &\sum_{i'\neq i}(h(\w_t,\x_{i'},y_{i'})-h(\w_t',\x_{i'},y_{i'}))^{2}+(|h(\w_t,\x_i,y_i)-h(\w_t',\x'_i,y'_i)|)^{2}\\
		\leq &(b-1)(\beta_{\w}\Delta\w_t)^{2} + (h(\w_t,\x_i,y_i)-h(\w_t',\x_i,y_i)\\
		&+h(\w_t',\x_i,y_i)-h(\x'_i,\w_t',y_i)+h(\x'_i,\w_t',y_i)-h(\w_t',\x'_i,y'_i))^{2}\\
		\leq &(b-1)(\beta_{\w}\Delta\w_t)^{2} + (\beta_{\w}\Delta\w_t+2\beta_{\x}+2\beta_yk_y)^{2}\\
		=&b(\beta_{\w}\Delta\w_t)^{2} +  2(2\beta_{\x}+2\beta_yk_y)\beta_{\w}\Delta\w_t + (2\beta_{\x}+2\beta_yk_y)^2.
		\end{align*}
		
		
		Then, consider multiple steps for the active party:
		\begin{align*}
		&[\Delta([\mathbf{IR}_t^A]_{t=1}^T)]^2\\
		&=\sum_{t=1}^{T}(\Delta([h(\x_i,\w_t,y_{i})]_{i\in \s_t}))^{2}
		\\&\leq (T-e)b(\beta_{\w}\Delta\w_T)^{2} + e(b(\beta_{\w}\Delta\w_T)^{2} +  2(2\beta_{\x}+2\beta_yk_y)\beta_{\w}\Delta\w_T + (2\beta_{\x}+2\beta_yk_y)^2)\\
		&\leq Tb(\beta_{\w}\Delta\w_T)^{2} +   e(2(2\beta_{\x}+2\beta_yk_y)\beta_{\w}\Delta\w_T + (2\beta_{\x}+2\beta_yk_y)^2)\\
		&\leq T\beta_{\w}^2\frac{4e^2\eta^2 L^2}{b} +   (2\beta_{\x}+2\beta_yk_y)\beta_{\w}\frac{4e^2\eta L}{b} + e(2\beta_{\x}+2\beta_yk_y)^2\\
		&= 4\beta_{\theta}^2L^2\frac{e^2T\eta^2 }{b} +   8(\beta_{\theta}k+\beta_yk_y)\beta_{\theta}L\frac{e^2\eta }{b} + 4(\beta_{\theta}k+\beta_yk_y)^2e\\
		&=[\Delta_2([\mathbf{IR}_t^A]_{t=1}^T)]^2,
		\end{align*}
		where for the second inequality, use the proof of lemma \ref{lemma:delta_w}.

	\end{proof}

	\subsection{Proof of Theorem \ref{th:dp-satisfy} [\xxx's Differential Privacy Guarantees]}
	\label{proof:dp-satisfy}
	\begin{proof}
		For the passive party, the view of the active party includes $[Sec[\mathbf{IR}_t^B]]_{t=1}^T$. Because we add perturbation to each element of the sequence by the $\ell_2$ sensitivity of $[\mathbf{IR}_t^B]_{t=1}^T$, i.e., $\Delta_2 ([\mathbf{IR}_t^B]_{t=1}^T)$ defined in Lemma~\ref{lemma:u-l2}, using the Gaussian Mechanism introduced in Lemma~\ref{th:gaussian} with the standard deviation of
		$\sqrt{2\log(1.25/\delta)}\frac{\Delta_2 ([\mathbf{IR}_t^B]_{t=1}^T)}{\epsilon}$, then by Lemma~\ref{th:gaussian} we have for all adjacent databases $D^B, {D'}^B$ that differ in a single data instance $\x_i^B\rightarrow {\x'_i}^{B}$, and for any set $\mathcal{S}\in \mathbb{R}^{T\times b}$, the Algorithm~\ref{alg:dp-vfl} satisfies:
		\begin{align*}
		\mathbb{P}([Sec[\mathbf{IR}_t^B]]_{t=1}^T\in \mathcal{S}|D^B) \leq \exp(\epsilon)\mathbb{P}([Sec[\mathbf{IR}_t^B]]_{t=1}^T\in \mathcal{S}|{D'}^B) + \delta.
		\end{align*}
		
		Similarly, for the active party, the view of the passive party includes $[Sec[\mathbf{IR}_t^A]]_{t=1}^T$. Because we add perturbation to each element of the sequence by the $\ell_2$ sensitivity of $[\mathbf{IR}_t^A]_{t=1}^T$, i.e., $\Delta_2 ([\mathbf{IR}_t^A]_{t=1}^T)$ defined in Lemma~\ref{lemma:v-l2}, using the Gaussian Mechanism introduced in Lemma~\ref{th:gaussian} with the standard deviation of
		$\sqrt{2\log(1.25/\delta)}\frac{\Delta_2 ([\mathbf{IR}_t^A]_{t=1}^T)}{\epsilon}$, then by Lemma~\ref{th:gaussian} we have for all adjacent databases $D^A, {D'}^A$ that differ in a single data instance $(\x_i^A,y_i^A)\rightarrow ({\x'_i}^{A},{y'_i}^A)$, and for any set $\mathcal{S}\in \mathbb{R}^{T\times b}$, the Algorithm~\ref{alg:dp-vfl} satisfies:
		\begin{align*}
		\mathbb{P}([Sec[\mathbf{IR}_t^A]]_{t=1}^T\in \mathcal{S}|D^A) \leq \exp(\epsilon)\mathbb{P}([Sec[\mathbf{IR}_t^A]]_{t=1}^T\in \mathcal{S}|{D'}^A) + \delta.
		\end{align*}
		
		Such properties can be easily demonstrated to hold for multiple passive parties.
		
		Then, according to the definition~\ref{df:dp}, Algorithm \ref{alg:dp-vfl} is $(\epsilon,\delta)$-differentially private w.r.t $[Sec[\mathbf{IR}_t^A]]_{t=1}^T$ and $[Sec[\mathbf{IR}_t^B]]_{t=1}^T$.
	\end{proof}
	
	\subsection{Proof of Theorem \ref{th:jdp-satisfy} [\xxx's Joint Differential Privacy Guarantees]}
	\label{proof:jdp-satisfy}
	\begin{proof}
		For the passive party, the view of the active party includes $[\w_t^A]_{t=1}^T$. Since the mapping $Sec[\textbf{IR}_t^B] \rightarrow \w_t^A$ does not touch any unperturbed sensitive information of $D^B$, the \emph{Post-Processing immunity property} (property \ref{property:post-processing}) can be applied such that combining the proof of Theorem \ref{th:dp-satisfy}, we have for all adjacent databases $D^B, {D'}^B$ that differ in a single data instance $\x_i^B\rightarrow {\x'_i}^{B}$, and for any set $\mathcal{S}\in \mathbb{R}^{T\times d}$, the Algorithm~\ref{alg:dp-vfl} satisfies:
		\begin{align*}
		\mathbb{P}([\w_t^A]_{t=1}^T\in \mathcal{S}|D^B,D^A) \leq \exp(\epsilon)\mathbb{P}([\w_t^A]_{t=1}^T\in \mathcal{S}|{D'}^B,D^A) + \delta.
		\end{align*}
		
		Similarly, for the active party, the view of the passive party includes $[\w_t^B]_{t=1}^T$. Since the mapping $Sec[\textbf{IR}_t^A] \rightarrow \w_t^B$ does not touch any unperturbed sensitive information of $D^A$, the \emph{Post-Processing immunity property} (property \ref{property:post-processing}) can be applied such that combining the proof of Theorem \ref{th:dp-satisfy}, we have for all adjacent databases $D^A, {D'}^A$ that differ in a single data instance $\x_i^A\rightarrow {\x'_i}^{A}$, and for any set $\mathcal{S}\in \mathbb{R}^{T\times d}$, the Algorithm~\ref{alg:dp-vfl} satisfies:
		\begin{align*}
		\mathbb{P}([\w_t^B]_{t=1}^T\in \mathcal{S}|D^A,D^B) \leq \exp(\epsilon)\mathbb{P}([\w_t^B]_{t=1}^T\in \mathcal{S}|{D'}^A,D^B) + \delta.
		\end{align*}
		
		Such properties can be easily demonstrated to hold for multiple passive parties.
		
		Then, according to the definition~\ref{df:jdp}, Algorithm \ref{alg:dp-vfl} is $(\epsilon,\delta)$-joint differentially private w.r.t $[\w_t^A]_{t=1}^T$ and $[\w_t^B]_{t=1}^T$.
	\end{proof}

	\subsection{Proof of Lemma \ref{lemma:error gradient} [Utility Analyses]}
	\label{proof:lm4}
	\begin{proof}
		
		In the $t$-th step of \xxx, the gradient error caused by the noisy data ($\mathbf{IR}$) is:
		\begin{align*}
		\|{\e}^{t}\|_2 = &\biggl\|\frac{1}{b}\sum_{i\in\s_t}\nabla \ell(\x_i\w_t,y_i)- [\mathbf{g}_t^A,\mathbf{g}_t^B]\biggr\|_2
		\\\leq &\frac{1}{b}\biggl\|\sum_{i\in\s_t}h(\x_i\w_t,y_i)\x_i- \sum_{i\in\s_t}\biggl(h(\x_i\w'_t + z_i^B,y_i)+ z_i^A\biggr)\x_i\biggr\|_2
		\\=&\frac{1}{b}\biggl\|\sum_{i\in\s_t} \x_i\biggl[h(\x_i\w_t,y_i)-h(\x_i\w'_t + z_i^B,y_i)- z_i^A\biggr]\biggr\|_2\\
		\leq&\frac{1}{b}\sum_{i\in\s_t} \biggl|h(\x_i\w_t,y_i)-h(\x_i\w'_t + z_i^B,y_i)- z_i^A\biggr|\\
		\leq&\frac{1}{b}\sum_{i\in\s_t} \biggl|h(\x_i\w_t,y_i)-h(\x_i\w'_t + z_i^B,y_i)\biggr|+ |z_i^A|\\
		\leq&\frac{1}{b}\sum_{i\in\s_t} \beta_{\theta}|\x_i\w_t - \x_i\w'_t - z_i^B|+ |z_i^A|\\
		\leq&\frac{1}{b}\sum_{i\in\s_t} \beta_{\theta}(\|\x_i\|\|\w_t-\w_t'\| + |z_i^B|)+ |z_i^A|\\
		\leq&\frac{1}{b}\sum_{i\in\s_t} \beta_{\theta}(2k + |z_i^B|)+ |z_i^A|.
		\end{align*}
		
		Because $z_i^A \sim \mathcal{N}(0,\sigma_A^2)$, $z_i^B \sim \mathcal{N}(0,\sigma_B^2)$, $\sigma_A = \sqrt{2\log(1.25/\delta)}\frac{\Delta_2 ([\mathbf{IR}_t^A]_{t=1}^T)}{\epsilon}$, and $\sigma_B = \sqrt{2\log(1.25/\delta)}\frac{\Delta_2 ([\mathbf{IR}_t^B]_{t=1}^T)}{\epsilon}$. According to tail inequality of Gaussian variable $z\sim \mathcal{N}(0,\sigma^2)$ such that $P(|z|\leq v)\geq 1-\frac{\sqrt{2}\sigma}{\sqrt{\pi}v}e^{-\frac{v^2}{2\sigma^2}}$ for $z>0$. Then for a constant $C>0$, with high probability of at least $1-\frac{\sqrt{2}}{\sqrt{\pi}C}e^{-\frac{C^2}{2}}$ we have: $|z| \leq C\sigma = O(\sigma)$,
		then we have:
		\begin{align*}
		\|\e^{t}\| =& O\biggl(\frac{\sqrt{2\log(1.25/\delta)}}{\epsilon}\biggl[\beta_{\theta}\biggl(2k+\sqrt{\frac{4L^2e^2T\eta^2}{b} + \frac{8k Le^2 \eta}{b} + 4k^2e}\biggr) \\
		&+ \sqrt{4\beta_{\theta}^2L^2\frac{e^2T\eta^2 }{b} +   8(\beta_{\theta}k+\beta_yk_y)\beta_{\theta}L\frac{e^2\eta }{b} + 4(\beta_{\theta}k+\beta_yk_y)^2e}\biggr]\biggr)
		\\=& O\biggl(\frac{\sqrt{2\log(1.25/\delta)}}{\epsilon}\biggl[ \sqrt{\frac{4\beta_{\theta}^2L^2e^2T\eta^2}{b} + \frac{8k \beta_{\theta}^2Le^2 \eta}{b} + 4\beta_{\theta}^2k^2e} \\
		&+ \sqrt{4\beta_{\theta}^2L^2\frac{e^2T\eta^2 }{b} +   8(\beta_{\theta}k+\beta_yk_y)\beta_{\theta}L\frac{e^2\eta }{b} + 4(\beta_{\theta}k+\beta_yk_y)^2e}\biggr]\biggr)
		\\=& O\biggl(\frac{\sqrt{2\log(1.25/\delta)}}{\epsilon}\sqrt{4\beta_{\theta}^2L^2\frac{e^2T\eta^2 }{b} +   8(\beta_{\theta}k+\beta_yk_y)\beta_{\theta}L\frac{e^2\eta }{b} + 4(\beta_{\theta}k+\beta_yk_y)^2e}\biggr)
		\\=& O\biggl(\frac{\sqrt{\log(1.25/\delta)}}{\epsilon}\sqrt{\frac{\beta_{\theta}^2L^2e^2T\eta^2 }{b} +   \frac{2(\beta_{\theta}k+\beta_yk_y)\beta_{\theta}Le^2\eta }{b} + (\beta_{\theta}k+\beta_yk_y)^2e}\biggr).
		\end{align*}
	\end{proof}
	\subsection{Proof of Theorem \ref{th:utility} [Utility Analyses]}
	\label{proof:th3}
	\begin{proof}
		Use Proposition 1 of \citet{schmidt2011convergence}, we have:
		\begin{align*}
		\calL\biggl(\frac{1}{T}\sum_{t=1}^{T}\w_{t}\biggr) - \calL(\w^{*}) &\leq \frac{\beta}{2T}\biggl(\|\w_0 - \w^{*}\| + 2\sum_{t = 1}^{T}\frac{\|\e^t\|}{\beta}\biggr)^2.
		\end{align*}
		Then we replace the gradient error in Proposition 1 with the $\|\e^{t}\|$ calculated in lemma \ref{lemma:error gradient}, and we get:
		\begin{align*}
		&\calL\biggl(\frac{1}{T}\sum_{t=1}^{T}\w_{i}\biggr) - \calL(\w^{*})=
		O\biggl(\biggl[k\sqrt{\frac{\beta}{T}} \\
		&+ 2\sqrt{\frac{T}{\beta}}\biggl(\frac{\sqrt{\log(1.25/\delta)}}{\epsilon}\sqrt{\frac{\beta_{\theta}^2L^2e^2T\eta^2 }{b} +   \frac{2(\beta_{\theta}k+\beta_yk_y)\beta_{\theta}Le^2\eta }{b} + (\beta_{\theta}k+\beta_yk_y)^2e}\biggr)\biggr]^{2}\biggr).
		\end{align*}
	\end{proof}

	
	%
	
	\section{Privacy-Accuracy Tradeoff}
	\label{appendix:exp}
\begin{figure}[ht]
		\begin{subfigure}{.33\textwidth}
			\centering
			\includegraphics[width=\linewidth]{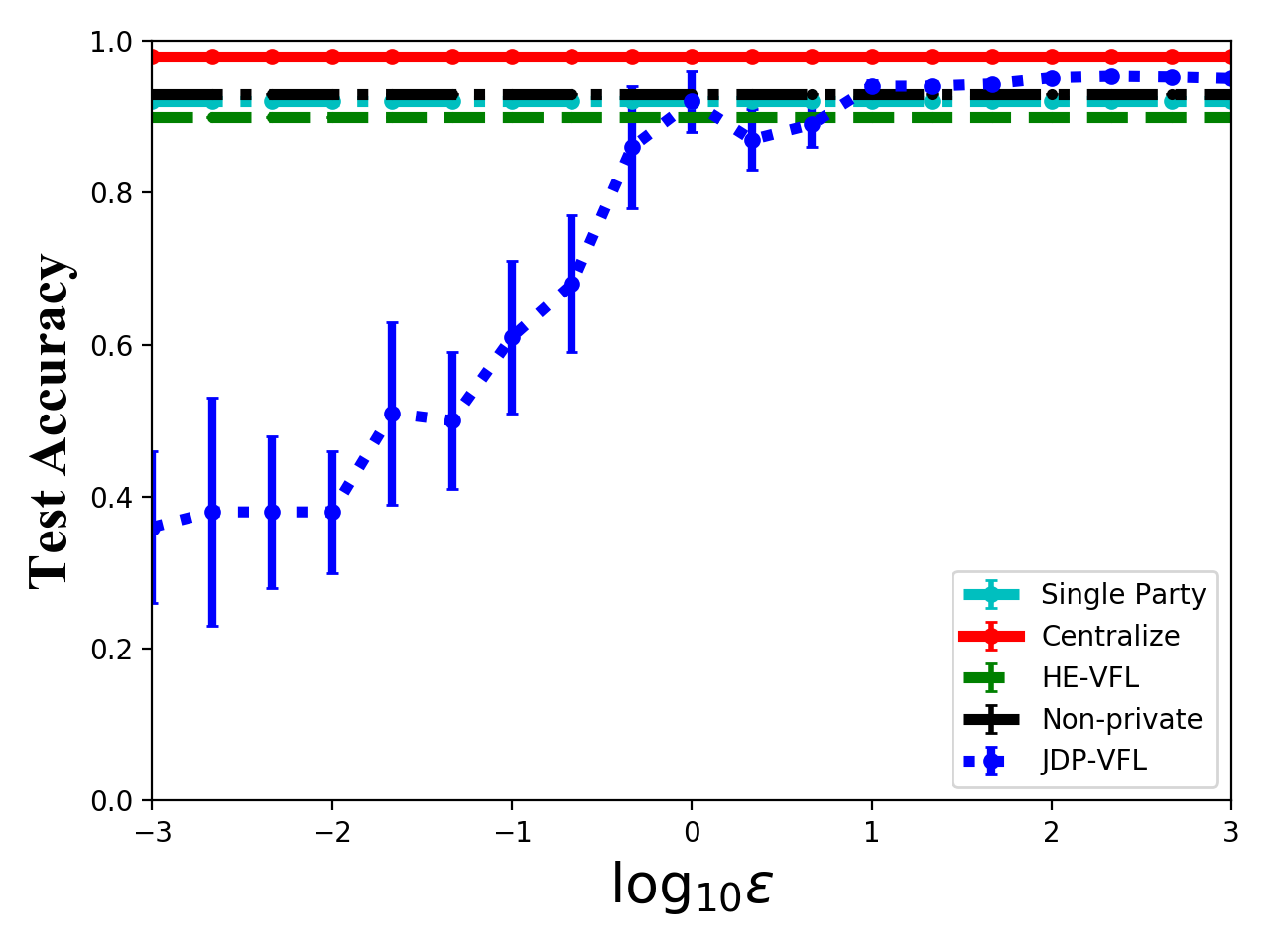}
			\caption{Breast}
		\end{subfigure}
		\begin{subfigure}{.33\textwidth}
			\centering
			\includegraphics[width=\linewidth]{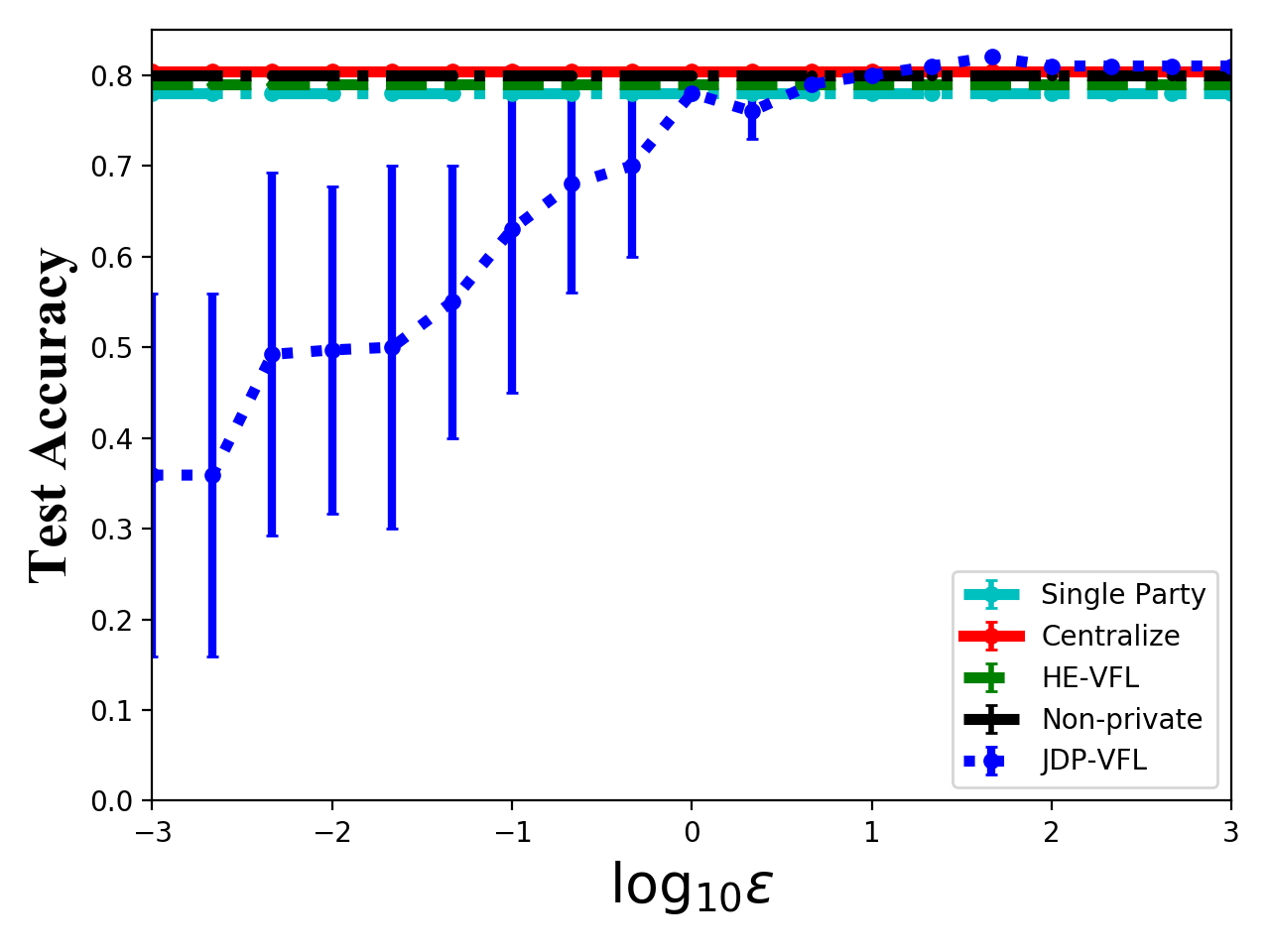}
			\caption{Credit}
		\end{subfigure}
		\begin{subfigure}{.33\textwidth}
			\centering
			\includegraphics[width=\linewidth]{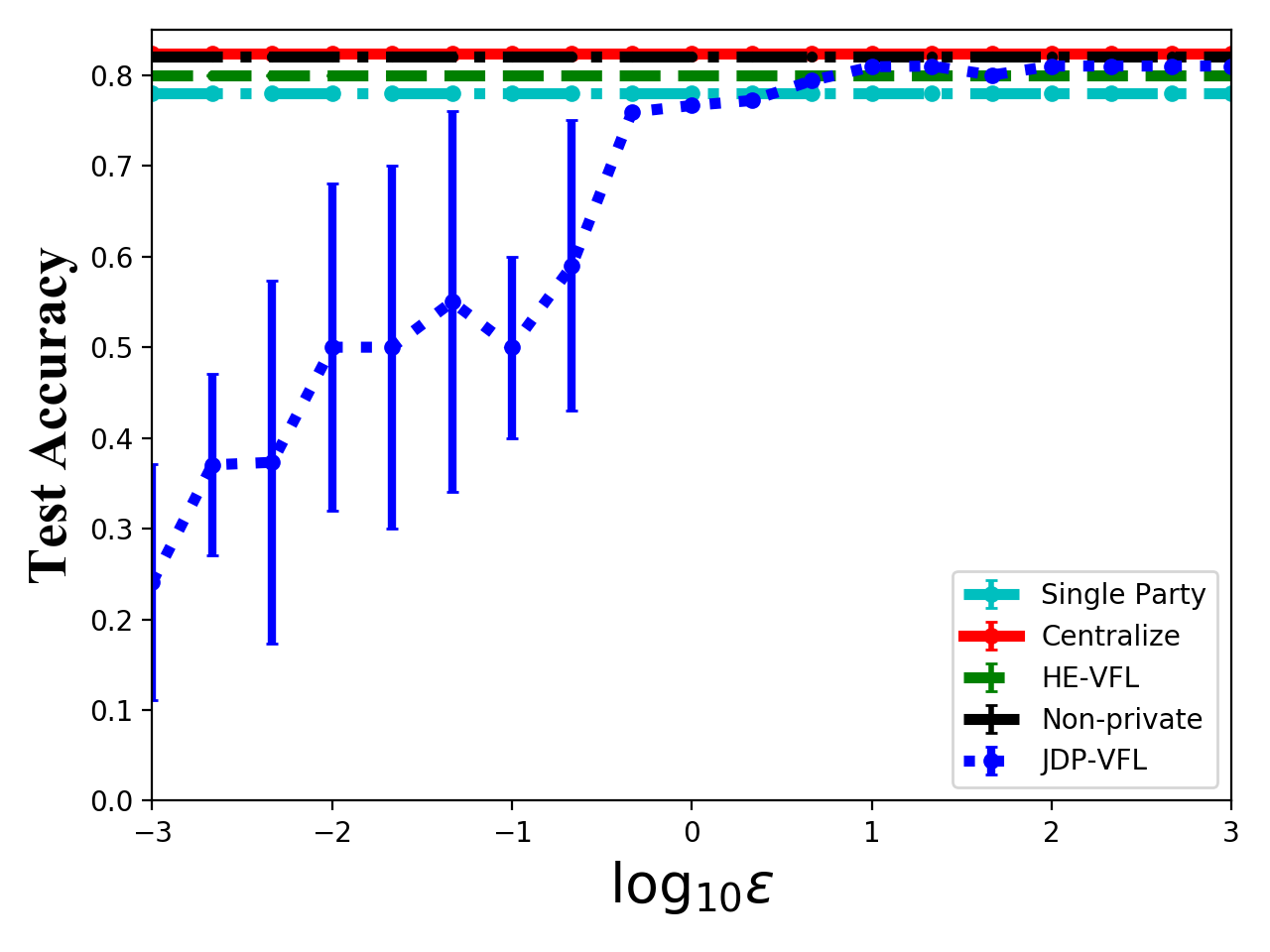}
			\caption{Adult}
		\end{subfigure}
		\caption{\xxx's privacy-accuracy tradeoff results using public dataset. We set mini-batch size $b = 3200$, $\lambda = 0.001$, epoch number $e = 10$, weight constraint $k = 1$.}
		\label{exp:whole trade off}
	\end{figure}

	In this section, we report the test-accuracy results on the full range of $\epsilon$ in $[0.001, 1000]$, as mentioned in Section \ref{sec:exp_setting}, in Figure\ref{exp:whole trade off}. From the results we can see that \xxx's accuracy is comparable to other evaluated methods if the privacy budget is sufficient, e.g., above $1$.
	
	\section{Extensions To Other Loss Functions and Penalties}\label{sec:extension}
	
	This section introduces additional loss functions and penalties which support the mainstream machine learning tasks. We show that \xxx~in Algorithm~\ref{alg:dp-vfl} can cover these commonly-encountered objective functions by merely changing some parameters, and the theoretical results still hold.
	
	\subsection{Extensions To Other Loss Functions}
	This section first introduces two additional losses for linear regression and classification, respectively, and then introduces losses for general applications, including Poisson regression and Gamma regression.

	\subsubsection{Least Square Loss}
	The least square loss is often used for linear regression which is widely applied for continuous-variable prediction. The loss function is as follows.
	\begin{align}
	\label{equ:least_square}
	\ell(\x_i\w,y_i) =(y_i-\x_i\w)^2, \ y_i\in\mathbb{R}.
	\end{align}
	
	Correspondingly, for $i\in\s_t$, each $i$-th entry of $\mathbf{IR}_t^A$ in Algorithm~\ref{alg:dp-vfl} equals
	\begin{align}
	\label{equ:vfl_gradient_agg_ls}
	\frac{\partial \ell}{\partial \theta_{i,t}}\biggl|_{\theta_{i,t} =\x_i\w_t^A+Sec[\mathbf{IR}_t^B]_i} = -2(y_{i}-\x_i\w_t^A-Sec[\mathbf{IR}_t^B]_i)). 
	\end{align}
	
	We can normalize the targets by subtracting the mean and dividing the standard deviance to approximate standard normal variables, then with high probability,
	other parameters are: $L=6,\beta_{\theta}=2,\beta_y=2,k_y=3$.
	
	\subsubsection{$\ell_2$-loss Support Vector Machine}
	The support vector machine is widely applied for classification. Enjoying smooth properties, the $\ell_2$-loss support vector machine is popular. The loss function is as follows.
	\begin{align}
	\label{equ:svm}
	\ell(\x_i\w,y_i) =(\max(0,1-y_i\x_i\w))^2, \ y_i\in\{-1,+1\}.
	\end{align}
	
	Correspondingly, for $i\in\s_t$, each $i$-th entry of $\mathbf{IR}_t^A$ in Algorithm~\ref{alg:dp-vfl} equals
	\begin{align}
	\label{equ:vfl_gradient_agg_svm}
	\frac{\partial \ell}{\partial \theta_{i,t}}\biggl|_{\theta_{i,t} =\x_i\w_t^A+Sec[\mathbf{IR}_t^B]_i} = -2y_{i}(\max(0,1-y_{i}(\x_i\w_t^A+Sec[\mathbf{IR}_t^B]_i))). 
	\end{align}
	
	Other parameters are: $L=2,\beta_{\theta}=2,\beta_y=2,k_y=1$.
	
	\subsubsection{Losses for The Exponential Dispersion Family}
	
	For general applications, this section introduce a type of loss function that follow a distribution from the exponential dispersion family~\cite{jorgensen1987exponential}:
	\begin{equation}
	\label{eq:glmmodel}
	\ell(y_{i};\theta_{i},\phi) = \frac{y_{i}\theta_{i}-b(\theta_{i})}{a(\phi)}+c(y_{i};\phi),
	\end{equation}
	where $\theta_{i}=\x_i\w$ is the natural parameter, $\phi$ is the dispersion parameter, and $a(\cdot)$, $b(\cdot)$, $c(\cdot)$ are known functions determined by the specific distribution, with some abuse of notation. This type of loss function covers a wide range of distribution, including Bernoulli, Normal, Poisson, and Gamma distributions for logistic regression, least square regression, Poisson regression, and Gamma regression, respectively. The specific forms of $a(\cdot)$, $b(\cdot)$, $c(\cdot)$ for these distributions are listed in Table~\ref{tab:dist}.
	
	Correspondingly, for $i\in\s_t$, each $i$-th entry of $\mathbf{IR}_t^A$ in Algorithm~\ref{alg:dp-vfl} equals
	\begin{align}
	\label{equ:vfl_gradient_agg_edf}
	\frac{\partial \ell}{\partial \theta_{i,t}}\biggl|_{\theta_{i,t} =\x_i\w_t^A+Sec[\mathbf{IR}_t^B]_i} = \frac{y_{i}-b'(\x_i\w_t^A+Sec[\mathbf{IR}_t^B]_i)}{a(\phi)}. 
	\end{align}
	
	Other parameters are: $L=k_y/a(\phi),\beta_{\theta}=\sup|b''(\cdot)|,\beta_y=1/a(\phi)$.
	
	\begin{table}
		\centering
		\caption{Some common distributions in the exponential dispersion family.}\label{tab:dist}
		\begin{tabular}{lccccc}
			\hline
			Distribution  & $\theta$ & $\phi$ & $a(\phi)$ & $b(\theta)$ & $c(y;\phi)$\\
			\hline
			\mbox{Bernoulli}($p$)  & $\log \{p(1-p)^{-1}\}$ & 1 & 1 & $\log(1+e^\theta)$ & 0 \\
			\mbox{Normal}($\mu$, $\sigma^2$)  & $\mu$ & $\sigma^2$ & $\phi$ & $\theta^2/2$ & $-(y^2\phi^{-1}+\log 2\pi)/2$ \\
			\mbox{Poisson}($\lambda$)  & $\log\lambda$ & 1 & 1 & $e^\theta$ & $-\log y!$\\
			\mbox{Gamma}($\alpha,\beta$)  & $-\beta/\alpha$ & $1/\alpha$ & $\phi$ & $ -\log(-\theta)$ & $\log(\alpha^\alpha y^{\alpha-1}/\Gamma(\alpha))$ \\
			\hline
		\end{tabular}
	\end{table}
	
	\subsection{Extensions To Other Penalties}
	
	This section introduces two popular penalties. Since these penalties result in element-wise operations which do not involve data instances, no additional privacy concern is required to address. Therefore, the privacy and utility bounds still hold.
	\subsubsection{$\ell_1$ Norm Penalty}
	
	$\ell_1$ norm penalty is popular to introduce sparseness into model weights for interpretation or information compression.
	For a $\ell_1$ norm penalty $\lambda\|\w\|_1$, one can update by proximal operators:
	\begin{align*}
	\mbox{Pen}(\w_t^{\cdot}, \g_t^{\cdot},\eta,\lambda)&= \mbox{sign}(\tilde{\w}_t^{\cdot})\max\{\mathbf{0},|\tilde{\w}_t^{\cdot}|-\eta\lambda\}\\
	\tilde{\w}_t^{\cdot}&=\w_t^{\cdot}-\eta\g_t^{\cdot}.
	\end{align*}
	
	\subsubsection{Elastic Net Penalty}
	
	Elastic net penalty is effective to achieve both sparseness and accurate estimation, which is a compromise between $\ell_1$ and $\ell_2$ norm regularization. 
	
	For an elastic net penalty norm penalty $\lambda[\|\w\|_1+(\mu/2)\|\w\|_2^2]$, one can also update by proximal operators:
	\begin{align*}
	\mbox{Pen}(\w_t^{\cdot}, \g_t^{\cdot},\eta,\lambda)&= \frac{1}{1+\eta\lambda\mu}\mbox{sign}(\tilde{\w}_t^{\cdot})\max\{\mathbf{0},|\tilde{\w}_t^{\cdot}|-\eta\lambda\}\\
	\tilde{\w}_t^{\cdot}&=\w_t^{\cdot}-\eta\g_t^{\cdot}.
	\end{align*}
	
\end{document}